%% file: log_det_jac.tex
\documentclass{article}

\pdfoutput=1

\usepackage[final, nonatbib]{neurips_2020}

\usepackage[utf8]{inputenc} % allow utf-8 input
\usepackage{comment}
\usepackage[T1]{fontenc}    % use 8-bit T1 fonts
\usepackage{hyperref}       % hyperlinks
\usepackage{url}            % simple URL typesetting
\usepackage{booktabs}       % professional-quality tables
\usepackage{amsfonts}       % blackboard math symbols
\usepackage{bm}             % bold mathematics
\usepackage{nicefrac}       % compact symbols for 1/2, etc.
\usepackage{microtype}      % microtypography
\usepackage{xcolor}
\usepackage{subcaption}
\usepackage[ruled,vlined]{algorithm2e}
\usepackage{amsmath}
\usepackage{kky}
\usepackage{enumitem}

\usepackage{graphicx}
\usepackage{pgfplots}
\pgfplotsset{compat=1.6}
\graphicspath{{figures/}}

\newcommand{\modified}[1]{{\color{black}  #1}}

\title{Relative gradient optimization of the Jacobian term\\ in 
unsupervised deep learning}

\author{
Luigi Gresele\footnotemark[1]\, $^{1,2}$ \quad\quad\quad 
Giancarlo Fissore\thanks{Equal contribution}\, $^{3,4}$ \quad\quad\quad Adri\'an Javaloy $^{1}$
\And Bernhard Sch\"olkopf $^{1}$\quad\quad\quad Aapo Hyv\"arinen $^{3, 5}$
\\[0.5em]
$^1$Max Planck Institute for Intelligent Systems, T\"ubingen, Germany\\
$^2$Max Planck Institute for Biological Cybernetics, T\"ubingen, Germany\\
$^3$ Universit\'e Paris-Saclay, Inria, Inria Saclay-\^{I}le-de-France, 91120, Palaiseau, France\\
$^4$ Universit\'e Paris-Saclay, CNRS, Laboratoire de recherche en informatique, 91405, Orsay, France\\
$^5$ Dept of Computer Science, University of Helsinki, Finland\\
\texttt{luigi.gresele@tuebingen.mpg.de; giancarlo.fissore@inria.fr}
}

\begin{document}

\maketitle

\begin{abstract}
    Learning expressive probabilistic models correctly describing the data 
    is a ubiquitous problem in machine learning.
    A popular approach for solving it is 
    mapping the observations into a representation space with a simple joint distribution, which can typically be written as a product of its marginals --- thus drawing a connection with the field of nonlinear independent component analysis. Deep density models have been widely used for this task, but their \modified{maximum} likelihood based training requires estimating the log-determinant of the Jacobian and is computationally expensive, thus imposing a trade-off between computation and expressive power. 
    In this work, we propose a new approach for exact %
    training of such neural networks. 
    Based on relative gradients, we exploit the matrix structure of neural network parameters to compute updates efficiently even in high-dimensional spaces; the computational cost of the training is quadratic in the input size, in contrast with the cubic scaling of naive approaches.
    This allows fast training with objective functions involving the log-determinant of the Jacobian, without imposing constraints on its structure, %
	in stark contrast to autoregressive normalizing flows.
\end{abstract}

\section{Introduction}
\label{sec:intro}
\input{sec/intro}

\section{Background}
\label{sec:logdetjac}

\input{sec/logdetjac}

\section{Log-determinant of the Jacobian for fully connected neural networks}
\label{sec:rel_grad}
\input{sec/rel_grad}

\section{Relative gradient descent for neural networks}
\label{sec:derivation}
\input{sec/derivation}

\paragraph{Jacobian term optimization through the relative gradient}

\input{sec/complexity_backprop}

\section{Experiments}
\label{sec:expts}

\input{sec/expts}

\section{Conclusions}
\label{sec:conclusions}

Using relative gradients, we proposed a new method for exact optimization of objective functions involving the log-determinant of the Jacobian of a neural network, as typically found in density estimation, nonlinear ICA, and related tasks. This allows for employing models which, unlike typical alternatives in the normalizing flows literature, have no strong limitation on the structure of the Jacobian. We use modules with fully connected layers, thus strictly generalizing normalizing flows with triangular Jacobians,  while still supporting efficient combination of forward and backward pass.
These neural networks can represent a larger function class than autoregressive flows, which, despite being universal approximators for density functions, can only represent transformations with triangular Jacobians.
Our method can therefore provide an alternative in settings where more expressiveness is needed to learn a proper inverse transformation, such as in identifiable nonlinear ICA models.

The relative gradient approach proposed here is quite simple, yet rather powerful. 
The importance of the optimization of the log-determinant of the Jacobian is well-known, but it has not been previously shown that there is a way around its difficulty without restricting expressivity. 
Now that we have shown that the optimization of this term can be done quite cheaply, a substantial fraction of the research in the field can be reformulated in stronger terms and with more generality. 

\section*{Broader impact}
As this paper presents novel theoretical results in unsupervised learning, the authors do not see any immediate ethical or societal concern. % 
An important aspect of our paper is the improvement in computational efficiency with respect to naive methods. This can hopefully lead to reduced energy consumption to achieve comparable model performance.

\section*{Acknowledgments}
A.H. was supported by a Fellowship from CIFAR, and by the DATAIA convergence institute as part of the "Programme d'Investissement d'Avenir", (ANR-17-CONV-0003) operated by Inria.\\ L.G. started working on this project while on an ELLIS exchange, hosted by the Parietal team at Inria, Saclay.\\
We thank Vincent Stimper, Patrick Putzky, Cyril Furtlehner and Ilyes Khemakhem for valuable comments on an earlier draft of this paper, and Isabel Valera and Guillaume Charpiat for helpful comments and tips. L.G. additionally thanks Roma Beaufret and Alexis Bozio for helpful support. %

\bibliographystyle{plain}
\bibliography{log_det_jac}

\clearpage

\appendix
\begin{center}
{\centering \LARGE APPENDIX}
\vspace{1cm}
\sloppy

\end{center}

\section{Backpropagation in neural networks}
\label{sec:mat_bp}

\input{app/backprop}
\section{Related work}
\label{app:app_rel_work}
\input{app/app_rel_work}

\section{Complexity of mathematical operations involved in gradient computation}
\label{app:complexity}

\input{app/complexity}

\section{Implementation details}
\label{app:bp_impl}
\input{app/implementation}

\section{Universal approximation capacity in normalizing flows}
\label{app:univ_approx}
\input{app/univ_approx}

\section{Relative gradient for the augmented matrix}
\label{app:augm_matrix}

\input{app/augm_matrix}

\section{Convolutions}
\label{app:convolutions}
\input{app/convolutions}
\section{Experiments}
\label{app:experiments}

\subsection{Computation of relative vs. ordinary gradient}
\input{app/computation_rel_vs_ord}

\subsection{Relative gradient optimization behaviour with different optimizers}
\input{app/optimization_rel}

\subsection{Density estimation}
\input{app/density}
\clearpage

\end{document}

%% file: sec/intro.tex
Many problems of machine learning and statistics involve learning invertible transformations of complex, multimodal probability distributions into simple ones%
. One example is density estimation through latent variable models under a specified base distribution~\cite{tabak2010density}, which can also have applications in data generation~\cite{dinh2014nice, kingma2018glow, grathwohl2018ffjord} and variational inference~\cite{rezende2015variational}. Another example is nonlinear independent component analysis (nonlinear ICA), where we want to extract simple, disentangled features out of the observed data~\cite{hyvarinen2016unsupervised, hyvarinen2018nonlinear}. 

One approach to learn such transformations, introduced in~\cite{tabak2013family} in the context of density estimation, is to represent them as a composition of simple maps, the sequential application of which enables high expressivity and a large class of representable transformations. Deep neural networks parameterize functions of multivariate variables as modular sequences of linear transformations and component-wise activation functions, thus providing a natural framework for implementing that idea, as already proposed in~\cite{rippel2013high}. 

Unfortunately, however, typical strategies employed in neural networks training do not scale well for objective functions like the aforementioned ones; in fact, through the change of variable formula, the logarithm of the absolute value of the determinant of the Jacobian appears in the objective. Its exact computation, let alone its %
optimization%
, quickly gets prohibitively computationally demanding as the data dimensionality grows. 

A large part of the research on deep density estimation, generally referred to under the term \emph{autoregressive normalizing flows}, has therefore been dedicated to considering a restricted class of transformations such that the computation of the Jacobian term is trivial~\cite{dinh2014nice, rezende2015variational, dinh2016density, kingma2016improved, huang2018neural, chen2019neural}, thus imposing a tradeoff between computation and expressive power. While such models can approximate arbitrary probability distributions, the extracted features are strongly restricted based on the imposed triangular structure, which prevents the system from learning a properly disentangled representation. Other strategies involve the optimization of an approximation of the exact objective~\cite{behrmann2018invertible}, and continuous-time analogs of normalizing flows for which the likelihood (or some approximation thereof) can be computed using relatively cheap operations~\cite{chen2018neural, grathwohl2018ffjord}.

In this work, we provide an efficient way to optimize the exact maximum likelihood objective for deep density estimation as well as for learning disentangled representations by latent variable models.
We consider a nonlinear, invertible transformation from the observed to the latent space which is parameterized through fully connected neural networks. The weight matrices are merely constrained to be invertible.
The starting point is that the parameters of the linear transformations are matrices; this allows us to exploit properties of the Riemannian geometry of matrix spaces to derive parameter updates in terms of the relative gradient, which was originally introduced as the natural gradient in the context of linear ICA~\cite{cardoso1996equivariant, amari1998natural}, and which can be feasibly computed. We show how this can be integrated with the usual backpropagation employed to compute gradients in neural network training, yielding an overall efficient way to optimize the Jacobian term in neural networks. This is a general optimization approach which is potentially useful for any objective involving such a Jacobian term, and is likely to find many applications in diverse areas of probabilistic modelling, for example in the context of Bayesian active learning for the computation of the information gain score~\cite{seeger2008large}, or for fitting the reverse Kullback-Leibler divergence in variational inference~\cite{wainwright2008graphical, blei2017variational}.

The computational cost of our proposed optimization procedure is quadratic in the input size---essentially the same as ordinary backpropagation--- which is  
in stark contrast with the cubic scaling of the naive way of optimizing via automatic differentiation. The joint asymptotic scaling of forward and backward pass as a function of the input size is therefore the same that aforementioned alternative methods achieve by imposing strong restrictions on the neural network structure~\cite{rezende2015variational} and thus on the class of functions they can represent. In contrast, our approach allows to efficiently optimize the exact objective for neural networks with arbitrary Jacobians.

In sections~\ref{sec:logdetjac} and~\ref{sec:rel_grad} we review maximum likelihood estimation for latent variable models, backpropagation and the Jacobian term for neural networks, and discuss the complexity of the naive approaches for optimizing the Jacobian term. Then in section~\ref{sec:derivation} we discuss the relative gradient, and show how it can be integrated with backpropagation resulting in an efficient procedure. 
We verify empirically the computational speedup our method provides in section~\ref{sec:expts}.

%% file: sec/logdetjac.tex
\subsection{Maximum likelihood for latent variable models}
\label{sec:density_est}
Consider a generative model of the form
\begin{equation}
\xb = \fb(\sbb)
\end{equation}
where $\sbb \in \mathbb{R}^D$ is the latent variable, $\xb\in \mathbb{R}^D$ represents the observed variable and $\fb : \mathbb{R}^D \rightarrow \mathbb{R}^D$ is a deterministic and invertible function, which we refer to as \emph{forward} transformation.  
Under the model specified above, the log-likelihood of a single datapoint $\xb$ can be written as
\begin{equation}
\label{eq:basic_likelihood1}
\log p_{\bm{\theta}}(\xb) = \log p_{s}(\gb_{\bm{\theta}}(\xb)) + \log |\det \Jb\gb_{\bm{\theta}}(\xb)|\,,
\end{equation}
where $\gb_{\bm{\theta}}$ is some representation with parameters $\bm{\theta}$ of the \emph{inverse} transformation\footnote{The forward transformation could also be parameterized, but here we only explicitly parameterize its inverse.} of $\fb$; \hbox{$\Jb\gb_{\bm{\theta}}(\xb) \in \mathbb{R}^{D \times D}$} its Jacobian computed at the point $\xb$, whose elements are the partial derivatives \hbox{$[\Jb\gb_{\bm{\theta}}(\xb)]_{ij}=\partial g^i_{\bm{\theta}}(\xb)/ \partial x^j$};  
and $p_{\bm{\theta}}$ and $p_{s}$ denote, respectively, the probability density functions of $\xb$ and of the latent variable $\sbb$ under the specified model. In many cases, it is additionally assumed that the distribution of the latent variable is sufficiently simple; for example, that it factorizes in its components,
\begin{equation}
\label{eq:basic_likelihood}
\log p_{\bm{\theta}}(\xb) = \sum_i \log p_i(\gb^i_{\bm{\theta}}(\xb)) + \log |\det \Jb\gb_{\bm{\theta}}(\xb)|\,.
\end{equation}
In this case, the problem can be interpreted as nonlinear independent component analysis (nonlinear ICA), and the components of $\gb_{\bm{\theta}}(\xb)$ are estimates of the original sources $\sbb$. 

Another variant of this framework can be developed to solve the problem that nonlinear ICA is, in general, not identifiable without additional assumptions~\cite{hyvarinen1999nonlinear}; that means, even if the data is generated according to the assumed model, there is no guarantee that the recovered sources bear any simple relationship to the true ones. In order to obtain identifiability, it is possible to consider models~\cite{hyvarinen2016unsupervised, hyvarinen2017nonlinear, hyvarinen2018nonlinear, gresele2019incomplete} in which the latent variables are not \emph{unconditionally} independent, but rather \emph{conditionally} independent given an additional, observed variable $\ub \in \mathbb{R}^d$,
\begin{equation}
\label{eq:cond_likelihood}
\log p_{\bm{\theta}}(\xb | \ub) = \sum_i \log p_i(\gb^i_{\bm{\theta}}(\xb)| \ub) + \log |\det \Jb\gb_{\bm{\theta}}(\xb)|\,,
\end{equation}
where $d$ can be equal to or different from $D$ depending on the model.

Maximum likelihood estimation for the model parameters amounts to finding, through optimization, the parameters $\bm{\theta}^*$ such that the expectation of the likelihood given by the expression in equation~\eqref{eq:basic_likelihood} is maximized. 
For all practical purposes, the expectation will be substituted with the sample average.
Specifically, for optimization purposes, we will be interested in the computation of a gradient of such term on mini-batches of one or few datapoints, such that stochastic gradient descent can be employed.

\subsection{Neural networks and backpropagation}
\label{sec:neur_backprop}

Neural networks provide a flexible parametric function class for representing $\gb_{\bm{\theta}}$ through a sequential composition of transformations, 
$\gb_{\bm{\theta}}=
\gb_{L} \circ \ldots \circ \gb_{2} \circ \gb_{1}\,,$
where $L$ defines the number of layers of the network. 
When an input pattern $\xb$ is presented to the network, it produces a final output $\zb_{L}$ and a series of intermediate outputs. 
By defining $\zb_0=\xb$ and $\zb_L=\gb_{\bm{\theta}}(\xb)$, we can write the forward evaluation as 
\begin{equation}
    \label{eq:fwd_pass}
    \zb_k = \gb_k(\zb_{k-1}) \enspace \text{for } k=1, \ldots, L\,.
\end{equation}

Each module $\gb_k$ of the network involves two transformations,
\begin{enumerate}[label=(\alph*)]
    \item a coupling layer $C_{\Wb_k}$, that couples the inputs to the layer with the parameters $\Wb_k$ to optimize;
    \item other arbitrary manipulations $\bm{\sigma}$ of inputs/outputs. Typically, these are element-wise nonlinear activation functions with fixed parameters; we can for simplicity think of them as operations of the form $\boldsymbol{\sigma}(\mathbf{x})=\left(\sigma\left(x_{1}\right), \ldots, \sigma\left(x_{n}\right)\right)$ applied to vector variables.
\end{enumerate}
The resulting transformation can thus be written as $\gb_k(\zb_{k-1})= \boldsymbol{\sigma}(C_{\Wb_k}(\zb_{k-1}))$.

We will focus on fully connected modules, where the coupling $C_{\Wb}$ is simply a matrix-vector multiplication between the weights $\Wb_k$ and the input to the $k$-th layer; overall, the transformation operated by such a module can be expressed as $\boldsymbol{\sigma}(\Wb_k \zb_{k-1})$. Another kind of coupling layer is given by convolutional layers, typically used in convolutional neural networks~\cite{lecun1989backpropagation}.

The parameters of the network are randomly initialized and then learned by gradient based optimization with an objective function $\Lcal$, which is a scalar function of the final output of the network. 
At each learning step, updates for the weights are proportional to the partial derivative of the loss with respect to each weight.

The computation of these derivatives is typically performed by backpropagation~\cite{rumelhart1986learning}, a specialized instance of automatic differentiation.
Backpropagation involves a two-phase process. Firstly, during a \emph{forward pass}, the intermediate and final outputs of the network $\zb_1, \ldots, \zb_L$ are evaluated and a value for the loss is returned.
Then, in a second phase termed \emph{backward pass}, derivatives of the loss with respect to each individual parameter of the network are computed by application of the chain rule. 
The gradients are computed one layer at a time, from the last layer to the first one; in the process, the intermediate outputs of the forward pass are reused, employing dynamic programming to avoid redundant calculations of intermediate, repeated terms.\footnote{Note that invertible neural networks provide the possibility to not save, but rather recompute the intermediate activations during the backward pass, thus providing a memory efficient approach to backpropagation~\cite{gomez2017reversible}.}

In matrix notation, the updates for the weights of the $k$-th fully connected layer $\Wb_k$ can then be written as  
\begin{equation}
    \Delta \Wb_k \propto  \zb_{k-1}  \bm{\delta}_{k}^{\top} \,, \label{eq:grad_upda}
\end{equation}
where $\bm{\delta}_{k}$ is the cumulative result of the backward computation in the backpropagation step up to the $k$-th layer, also called backpropagated error. We report the full derivation in appendix~\ref{sec:mat_bp}.
We adopt the convention of defining $\xb$, $\zb_k$ and $\bm{\delta}_{k}$ as column vectors.

\subsection{Difficulty of optimizing the Jacobian term of neural networks}

\label{sec:difficultyof}

In the case of the objective function specified in Eq.~(\ref{eq:basic_likelihood}), we have $\Lcal(\xb) = \log p_{\bm{\theta}}(\xb)$. By defining
\begin{equation}
    \Lcal_p (\xb) = \sum_i \log p_i(\gb^i_{\bm{\theta}}(\xb)) ; \enspace \ \Lcal_J (\xb) = \log \left|\det \Jb\gb_{\bm{\theta}}(\xb)\right|\,, 
\end{equation}
the objective can be rewritten as $\Lcal(\xb) = \Lcal_p(\xb) + \Lcal_J(\xb)$.
The evaluation of the gradient of the first term $\Lcal_p$
can be performed easily if a simple form for the latent density is chosen, as it only requires simple operations on top of a single forward pass of the neural network. Given that the loss is a scalar, as backpropagation is an instance of reverse mode differentiation~\cite{baydin2018automatic}, backpropagating the error relative to it in order to evaluate the gradients does not increase the overall complexity with respect to the forward pass alone.

In contrast, the evaluation of the gradient of the second term, $\Lcal_J$, is very problematic, and our main concern in this paper. 
The key computational bottleneck is in fact given by the evaluation of the Jacobian during the forward pass. Since the Jacobian involves derivatives of the function $\gb_{\bm{\theta}}$ with respect to its inputs $\xb$, this evaluation can again be performed through automatic differentiation. 
Overall, it can be shown~\cite{baydin2018automatic} that both forward and backward mode automatic differentiation for a $L$-layer, fully connected neural network scale as $\mathcal{O}( LD^3)$, with $L$ the number of layers. This is prohibitive in many practical applications with a large data dimension $D$.

\textbf{Normalizing flows with simple Jacobians}  An approach to alleviate the computational cost of this operation 
is to deploy special neural network architectures for which the evaluation of $\Lcal_J$ is trivial. 
For example, in autoregressive normalizing flows~\cite{dinh2014nice, dinh2016density, kingma2016improved, huang2018neural} the Jacobian of the 
transformation is constrained to be lower triangular. 
In this case, its determinant can be trivially computed with a linear cost in $D$. 
Notice however that the computational cost of the forward pass still scales quadratically in $D$; 
the overall complexity of forward plus backward pass is therefore still quadratic in the input size~\cite{rezende2015variational}. 

Most critically, such architectures imply a strong restriction on the class of transformations that can be learned. While it can be shown, based on~\cite{hyvarinen1999nonlinear}, that under certain conditions this class of functions has universal approximation capacity for \emph{densities}~\cite{huang2018neural}, that is less general than other notions of universal approximation~\cite{hornik1989multilayer, hornik1991approximation}. In fact it is obvious that functions with such triangular Jacobians cannot be universal approximators of \emph{functions}, since, for example, the first variable can only depend on the first variable. This is a severe problem in learning features for disentanglement, for example by nonlinear ICA~\cite{hyvarinen2016unsupervised, hyvarinen2018nonlinear},  
which would usually require unconstrained Jacobians. 
In other words, such restrictions might imply that the deployed networks are not general purpose: \cite{behrmann2018invertible} showed that constrained designs typically used for density estimation can severely hurt discriminative performance. We further elaborate on this point in appendix~\ref{app:univ_approx}.
Note that fully connected modules have elsewhere been termed \emph{linear} flows~\cite{papamakarios2019normalizing}, and are a strict generalization of autoregressive flows.\footnote{Comprehensive reviews on normalizing flows can be found in~\cite{papamakarios2019normalizing, kobyzev2019normalizing}. Other related methods are reviewed in appendix~\ref{app:app_rel_work}.}

%% file: sec/rel_grad.tex
As a first step toward efficient optimization of the $\Lcal_J$ term, we next provide the explicit form of the Jacobian for fully connected neural networks.
As a starting point, notice that invertible and differentiable transformations are \emph{composable}; given any two such transformations, their composition is also invertible and differentiable. Furthermore, the determinant of the Jacobian of a composition of functions is given by the product of the determinants of the Jacobians of each function,
\begin{align}
\operatorname{det} \Jb[\gb_2 \circ \gb_1] (\mathbf{x}) &=\operatorname{det} \Jb\gb_2\left( \gb_{1}(\mathbf{x})\right) \cdot \operatorname{det} \Jb\gb_1(\mathbf{x})\,. \end{align}
The log-determinant of the full Jacobian for a neural network therefore simply decomposes in a sum of the log-determinants of the Jacobians of each module, $\Lcal_J (\xb) = \sum_{k=1}^L \log |\operatorname{det} \Jb\gb_k(\mathbf{z}_{k-1})|  $.
We will focus on the Jacobian term relative to a single submodule $k$ with respect to its input $\zb_{k-1}$; with a slight abuse of notation, we will call it $\Lcal_J (\zb_{k-1})$.
As we remarked, fully connected $\gb_k$ are themselves compositions of a linear operation and an element-wise invertible nonlinearity; 
applying the same reasoning, we then have %
\begin{equation}
    \Lcal_J (\zb_{k-1}) 
    = \sum_{i=1}^D\log \left| \sigma'(y_{k}^{i}) \right| + \log \left| \det \Wb_k \right| \\
    =: \Lcal_J^{1}(\yb_{k}) + \Lcal_J^{2}(\zb_{k-1})\,.
\end{equation}%
where $\yb_k = \Wb_k \zb_{k-1}$.
The first term $\Lcal_J^{1}$ is a sum of univariate functions of single components of the output of the module, and it can be evaluated easily with few additional operations on top of intermediate outputs of a forward pass; gradients with respect to it can be simply computed via backpropagation, not unlike the $\Lcal_{p}$ term introduced in section~\ref{sec:difficultyof}. 

The second term $\Lcal_J^{2}$ however involves a nonlinear function of the determinant of the weight matrix. From matrix calculus, we know that the derivative is equal to
\begin{equation}
\frac{\partial \log |\operatorname{det} \mathbf{W}_k|}{\partial \mathbf{W}_k}=
\left(\mathbf{W}_k^{\top}\right)^{-1}\,.
\label{eq:mat_inv}
\end{equation}
Therefore, the computation of the gradient relative to such term involves a matrix inversion, with cubic scaling in the input size.\footnote{Though slightly more favorable exponents can in principle be obtained, see appendix~\ref{app:complexity}.} For a fully connected neural network of $L$ layers, given that we have one such operation to perform for each of the layers, the gradient computation for these terms alone would have a complexity of $\mathcal{O}(L D^3)$, thus matching the one which would be obtained if the Jacobian were to be computed via automatic differentiation as discussed in section~\ref{sec:logdetjac}. 

It can therefore be seen that these inverses of the weight matrices are the problematic element in the gradient computation.
In the next section, we show how this problem can be solved using relative gradients.

%% file: sec/derivation.tex
We now derive the basic form of the relative gradient, following the approach in \cite{cardoso1996equivariant}.\footnote{For linear blind source separation, this approach also corresponds to the natural gradient, which can be justified with an information-geometric approach~\cite{amari1998natural}.
}
The starting point is that the parameters in a neural networks are matrices, in particular invertible in our case. Thus, we can make use of the geometric properties of invertible matrices, while they are usually completely neglected in gradient optimization in neural networks.  %

\paragraph {Relative gradient based on multiplicative perturbation}
In a classical gradient approach for optimization, we add a small vector $\boldsymbol{\epsilon}$ to a point $\xb$ in a Euclidean space. However, with matrices, we are actually perturbing a matrix with another, and this can be done in different ways. In the relative gradient approach, we make a \textit{multiplicative} perturbation of the form
\begin{equation}
    \Wb_k \rightarrow (\Ib + \epsilonb) \Wb_k
\end{equation}
where $\epsilonb$ is an infinitesimal matrix. If we consider the effect of such a perturbation on a scalar-valued function $f(\Wb_k)$, we have
\begin{equation}
    f((\Ib+\epsilonb)\Wb_k)-f(\Wb)=\langle \nabla f(\Wb_k) , \epsilonb \Wb_k \rangle + o(\Wb_k) = \langle \nabla f(\Wb_k) \Wb_k^{\top}, \epsilonb \rangle + o(\Wb_k)  
\end{equation}
which shows that the direction of steepest descent in this case is given by making $\epsilonb= \mu \nabla f(\Wb_k) \Wb_k^{\top}$ where $\mu$ is an infinitesimal step size. Furthermore, when we combine this $\epsilonb$ with the definition of a multiplicative update, we find that the best perturbation to $\Wb$ is actually given as 
\begin{equation}
    \Wb_k \rightarrow \Wb_k + \mu \nabla f(\Wb_k) \Wb_k^{\top} \Wb_k
\end{equation}
That is, the classical Euclidean gradient is replaced by $\nabla f(\Wb_k) \Wb_k^{\top} \Wb_k$, i.e.\ it is multiplied by $\Wb_k^{\top}\Wb_k$ from the right. This is the relative gradient.

A further alternative can be obtained by perturbing the weight matrices from the right, as
\hbox{$\Wb_k \rightarrow \Wb_k (\Ib + \epsilonb)$}.
A similar derivation shows that in this case, the optimal $\epsilonb$ is given by $\Wb_k\Wb_k^{\top}\nabla f(\Wb_k)$; we refer to this as \emph{transposed relative gradient}. In the context of linear ICA, the properties of the relative and transposed relative gradient were discussed in~\cite{squartini2005new}.
This version of the relative gradient might be useful in some cases; for example, the transposed relative gradient can be implemented more straightforwardly in neural network packages where the convention is that vectors are represented as rows.

The relative gradient belongs to the more general class of gradient descent algorithms on Riemannian manifolds~\cite{absil2009optimization}. Specifically, relative gradient descent is a first order optimization algorithm on the manifold of invertible $D \times D$ matrices.  
Almost sure convergence of the parameters to a critical point of the gradient of the cost function can be 
derived even for its stochastic counterpart, with decreasing step size and under suitable assumptions (see e.g.~\cite{bonnabel2013stochastic}).%

%% file: sec/complexity_backprop.tex
In section~\ref{sec:rel_grad}, we showed that the difficulty in computing the gradient of the log-determinant is in the terms $\Lcal_J^{2}$, whose gradient involves a matrix inversion. Now we show that by exploiting the relative gradient, this matrix inversion vanishes. In fact, when multiplying the right hand side of equation~\eqref{eq:mat_inv} by $\Wb_k^{\top}\Wb_k$ from the right we get 
\begin{equation}
    \left(\mathbf{W}_k^{\top}\right)^{-1} \Wb_k^{\top}\Wb_k = \Wb_k \,,
\end{equation}
and similarly when multiplying by $\Wb_k\Wb_k^{\top}$ from the left.
Most notably, we therefore have to perform \emph{no additional operation} to get the relative gradient with respect to this term of the loss; it is, so to say, \emph{implicitly} computed --- as we know that the update for the parameters in $\Wb_k$ with respect to the error term $\Lcal_{J}^2$ is proportional to $\Wb_k$ matrix itself. 

As for the remaining terms of the loss, $\Lcal_p$ and $\Lcal^1_{J}$, 
simple backpropagation allows us to compute the weight updates given by the ordinary gradient in equation~\eqref{eq:grad_upda}, which still need to be multiplied by $\Wb_k^{\top} \Wb_k$ to turn it into a relative gradient. We will next see that we can do this avoiding matrix-matrix multiplications, which would be computationally expensive. 
Note that backpropagation necessarily computes the $\bm{\delta}_k$ vector in equation~\eqref{eq:grad_upda} and for our model, 
by applying the relative gradient carefully, we can avoid matrix-matrix multiplication altogether by computing
\begin{equation}
    \left(\Delta \Wb_k\right) \Wb_k^{\top} \Wb_k \propto \zb_{k-1} \left( \left(\bm{\delta}_k^{\top} \Wb_k^{\top} \right) \Wb_k \right)\,. \label{eq:grad_upd}
\end{equation}
Thus, we have a cheap method for computing the gradient of the log-determinant of the Jacobian, and of our original objective function. In appendix~\ref{app:bp_impl} we provide an explanation of how our procedure can be implemented with relative ease on top of existing deep learning packages.

While we so far only discussed update rules for the weight matrices of the neural network, our approach can be extended to include biases. 
Including bias terms in our multilayer network endows it with stronger approximation capacity. We detail how to do this in appendix~\ref{app:augm_matrix}.

\textbf{Complexity} Note that the parentheses in equation~\eqref{eq:grad_upd} stress the point that the relative gradient updates 
only require matrix-vector or vector-vector multiplications, each of which scales as $\mathcal{O}(D^2)$, in a fixed number at each layer; that is, overall $\mathcal{O}(L D^2)$ operations. They therefore do not increase the complexity of a normal forward pass. Furthermore, the overall complexity with respect to the input size is quadratic, resulting in an overall quadratic scaling with the input size as in normalizing flow methods~\cite{rezende2015variational}, but without imposing strong restrictions on the Jacobian of the transformation.

\textbf{Extension to convolutional layers} As we remarked in section~\ref{sec:neur_backprop}, the formalism we introduced includes convolutional neural networks (CNNs)~\cite{lecun1989backpropagation}. A natural question is therefore whether our approach can be extended to that case. The first natural question pertains the invertibility of convolutional neural networks; the convolution operation was shown~\cite{ma2018invertibility} to be invertible under mild conditions (see appendix~\ref{app:convolutions}), and the standard pooling operation can be by replaced an invertible operation~\cite{jacobsen2018revnet}. We therefore believe that the general formalism can be applied to CNNs; this would require the derivation of the relative gradient for tensors. We believe that this should be possible but leave it for future work.

\textbf{Invertibility and generation} 
Given that invertible and differentiable transformations are composable, as discussed in section~\ref{sec:rel_grad}, invertibility of our learned transformation is guaranteed as long as the weight matrices and the element-wise nonlinearities are invertible. Square and randomly initialized (e.g. with uniform or normally distributed entries) weight matrices are known to be invertible with probability one; invertibility of the weight matrices throughout the training is guaranteed by the fact that the $\Lcal_J^{2}$ terms would go to minus infinity for singular matrices (though high learning rates and numerical instabilities might compromise it in practice), as in estimation methods for linear ICA~\cite{bell1995information, cardoso1996equivariant, hyvarinen1999fast}.
We additionally employ nonlinearities which are invertible by construction; we include more details about this in appendix~\ref{app:experiments}.
If we are interested in data generation, we also need to invert the learned function. In practice, the cost of inverting each of the matrices is $\mathcal{O}(D^3)$, but the operation needs to be performed only once. As for the nonlinear transformation, the inversion is cheap since we only need to numerically invert a scalar function, for which often a closed form is available.

%% file: sec/expts.tex
\begin{figure}
  \centering
  \begin{subfigure}{.49\linewidth}
    \includegraphics[width=1.05\linewidth]{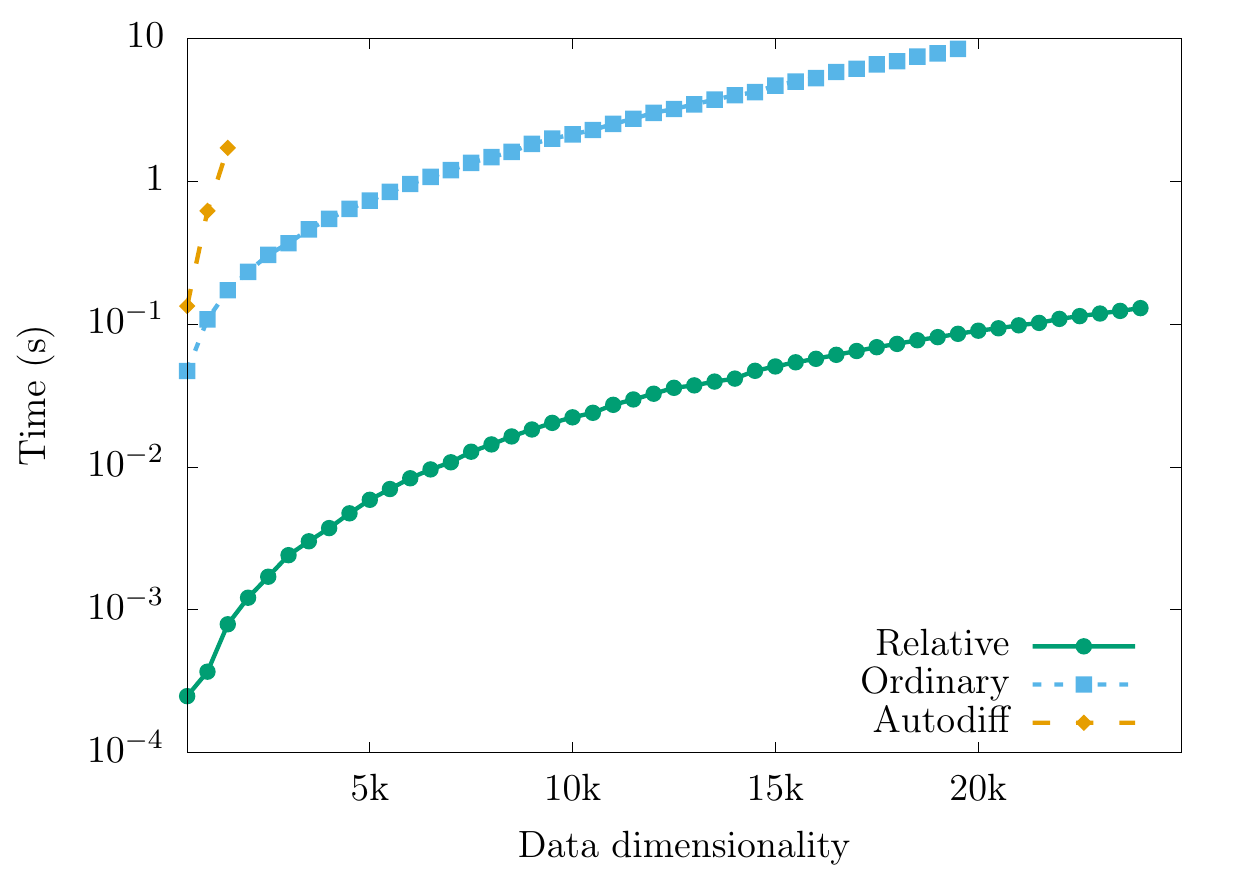}
    % \caption{\label{fig:complexity}}
  \end{subfigure}
  \begin{subfigure}{.49\linewidth}
    \includegraphics[width=1.05\linewidth]{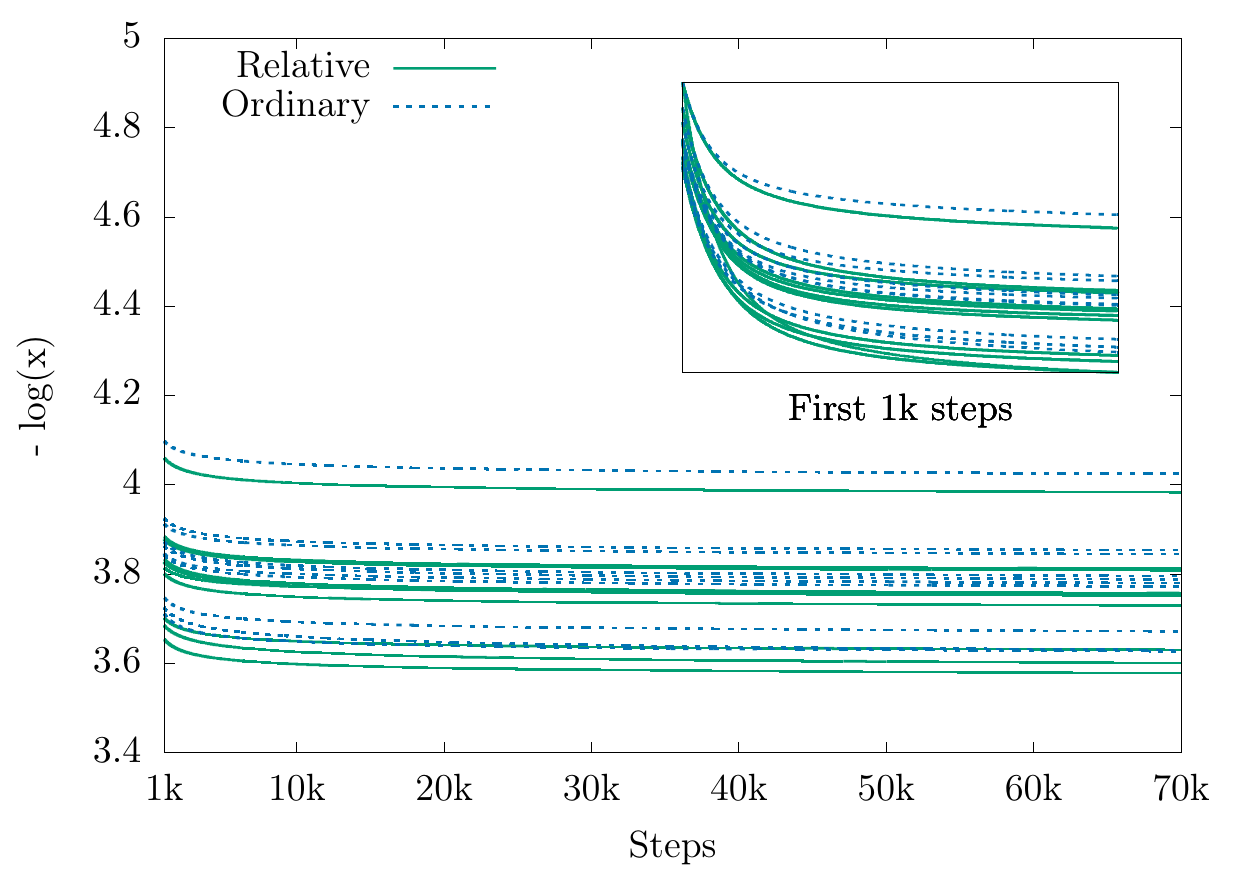}
    % \caption{\label{fig:loss_and_grads}}
  \end{subfigure}
    \caption{\textbf{Left: }
    Comparison of the average computation times of a single evaluation of the gradient of the log-likelihood;
    the standard error of the mean is not reported as it is orders of magnitude smaller then the scale of the plot.
    \textbf{Right: } Time-evolution of the negative log-likelihood for deterministic full-batch optimization for the two methods with the same initial points.\label{fig:comparison}}
\end{figure}

In the following we experimentally verify the computational advantage of the relative gradient. The code used for our experiments can be found at \url{https://github.com/fissoreg/relative-gradient-jacobian}.

\textbf{Computation of relative vs.\ ordinary gradient}
As a first step, we empirically verify that our proposed procedure using the formulas in section~\ref{sec:derivation} leads to a significant speed-up in
computation of the gradient of the Jacobian term. We compare the relative gradient against an explicit computation of the ordinary gradient, as described in section~\ref{sec:rel_grad}, and with a computation based on automatic differentiation, as discussed in section~\ref{sec:difficultyof}, where the Jacobian is computed with the JAX package~\cite{jax2018github}. While the output and asymptotic computational complexity of the ordinary gradient and automatic differentiation methods should be the same, a discrepancy is to be expected at finite dimensionality due to differences in how the computation is implemented.
In the experiment, we generate 100 random normally distributed datapoints and vary the dimensionality of the data from 10 to beyond 20,000. We then define a two-layer neural network and evaluate the gradient of the Jacobian.
The main comparison is run on a Tesla P100 Nvidia GPU. For the main plots, we deactivated garbage collection. Plots with CPU and further details on garbage collection can be found in appendix~\ref{app:exp_computaion}. For each dimension we computed 10 iterations with a batch size of 100.
Results are shown in figure \ref{fig:comparison}, left. On the y-axis 
we report the average of the execution times of 100 successive gradient evaluations (forward plus backward pass in the automatic differentiation case). 
It can be clearly seen that \emph{the relative gradient is much faster}, typically by two orders of magnitude. 
Autodiff computations could actually only be performed for the smallest dimension due to a memory problem. We report additional details on memory consumption in appendix~\ref{app:exp_computaion}. %

\textbf{Optimization by relative vs.\ ordinary gradient}
Since our paper is, to the best of our knowledge, the first one proposing relative gradient optimization for neural networks (though other kinds of natural gradients have been studied \cite{amari1998natural}), we want to verify that the learning dynamics induced by the relative as opposed to the ordinary gradient do not bias the training procedure towards less optimal solutions or create other problems.
We therefore perform a deterministic (full batch) gradient descent for both the relative and the ordinary gradient.\footnote{Notice that there's no need to compare to autodiff in this case because the computed gradient should be exactly the same as the ordinary  gradient with the formulas in section~\ref{sec:rel_grad}.} We employ 1,000 datapoints of dimensionality 2 
and a two-layer neural network. We take 10 initial points and initialize both kinds of gradient descent at those same points. On the x-axis we plot the training epoch, while on the y-axis we plot the value of the loss. 
Figure \ref{fig:comparison}, right shows the results: there is no big difference between the two gradient methods. There may actually be a slight advantage for the relative gradient, but that is immaterial since our main point here is merely to show that the \emph{relative gradient does not need more iterations} to give the same performance. 

Combining these two results, we see that the proposed relative gradient approach leads to a \emph{much faster optimization} than the ordinary gradient. Perhaps surprisingly, the results exhibit a rather constant speed-up factor of the order of 100 although the theory says it should be changing with the dimension $D$; in any case, the difference is very significant in practice.

\paragraph{Density estimation}
\begin{figure}[t]
    \centering
        \begin{minipage}{.164\textwidth}
            \begin{subfigure}{1.0\textwidth}
            \centering
            \includegraphics[height=2.29cm, keepaspectratio]{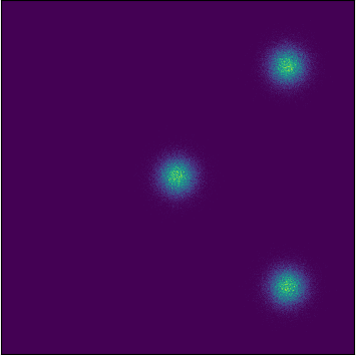}
            \end{subfigure}
        \end{minipage}%
        \begin{minipage}{.164\textwidth}
            \begin{subfigure}{1.0\textwidth}
            \centering
            \includegraphics[height=2.29cm, keepaspectratio]{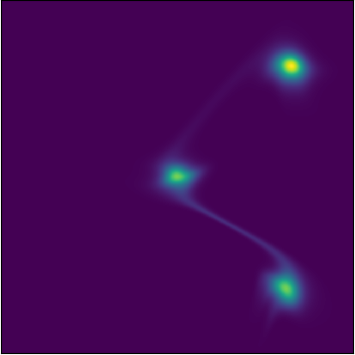}
            \end{subfigure}
        \end{minipage}%
        \hspace{.05em}
        \begin{minipage}{.164\textwidth}
            \begin{subfigure}{1.0\textwidth}
            \centering
            \includegraphics[height=2.29cm, keepaspectratio]{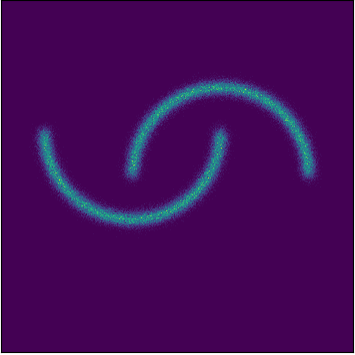}
            \end{subfigure}
        \end{minipage}%
        \begin{minipage}{.164\textwidth}
            \begin{subfigure}{1.0\textwidth}
            \centering
            \includegraphics[height=2.29cm, keepaspectratio]{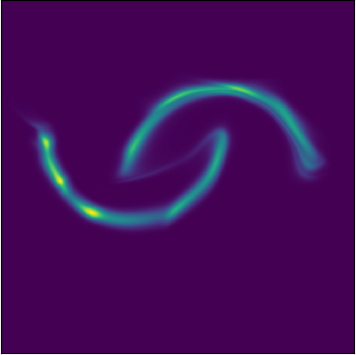}
            \end{subfigure}
        \end{minipage}%
        \hspace{.05em}
        \begin{minipage}{.164\textwidth}
            \begin{subfigure}{1.0\textwidth}
            \centering
            \includegraphics[height=2.29cm, keepaspectratio]{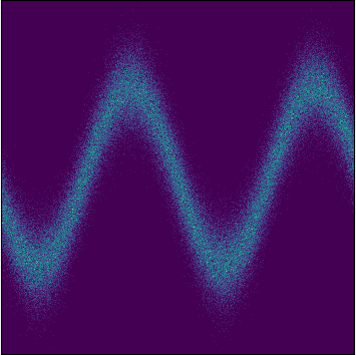}
            \end{subfigure}
        \end{minipage}%
        \begin{minipage}{.164\textwidth}
            \begin{subfigure}{1.0\textwidth}
            \centering
            \includegraphics[height=2.29cm, keepaspectratio]{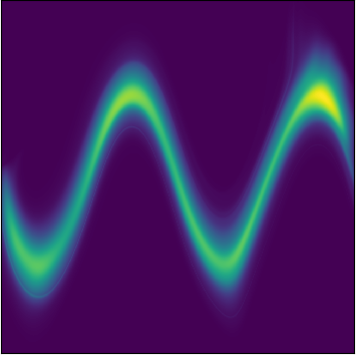}
            \end{subfigure}
        \end{minipage}%
        \caption{Illustrative examples of 2D density estimation. Samples from the true distribution and predicted densities are shown, in this order, side by side.}
        \label{fig:toy_density}
\end{figure}
Although our main contribution is the computational speed-up of the gradient computation demonstrated above, we further show some simple results on density estimation to highlight the potential of the relative gradient used in conjuction with the unconstrained factorial approximation in section~\ref{sec:density_est}.
We use a fairly simple feedforward neural network with a smooth version of leaky-ReLU as activation function. Our empirical results show that this system, despite having quite \emph{minimal fine-tuning} (details in appendix~\ref{app:exp_density_est}), \emph{achieves competitive results on all the considered datasets} compared with existing models---which are all tailored and fine-tuned for density estimation. 
First, we show in Figure~\ref{fig:toy_density} different toy examples that showcase the ability of our method to convincingly model arbitrarily complex densities.
Second, in order to show the viability of our method in comparison with well-established methods we perform, as in~\cite{papamakarios2017masked}, unconditional density estimation on four different UCI datasets~\cite{Dua:2019} and a dataset of natural image patches (BSDS300)~\cite{martin2001database}, as well as on MNIST~\cite{lecun1998mnist}. The results are shown in Table~\ref{tab:results}.
To achieve a fair comparison across models, the number of parameters was tuned so that the number of trainable parameters are as similar as possible.
Note that, as we can perform every computation efficiently, all the experiments are suitable to run on usual hardware, thus avoiding the need of hardware accelerators such as GPUs.
As a final remark, the reported results make no use of batch normalization, dropout, or learning-rate scheduling. Therefore, it is sensible to expect even better results by including them in future work.
\begin{table}[ht]
	\centering
	\caption{Test log-likelihoods (higher is better) on unconditional density estimation for different datasets and models (same as in Table 1 of~\cite{papamakarios2017masked}). Models use a similar number of parameters; results show mean and two standard deviations. Best performing models are in bold. More details in appendix~\ref{app:exp_density_est}} 
\label{tab:results}
	\resizebox{\textwidth}{!}{
    	\begin{tabular}{lcccccc} \toprule
    		 & POWER & GAS & HEPMASS & MINIBOONE & BSDS300 & MNIST \\ \midrule
    		Ours & $0.065\pm0.013$ & $6.978\pm0.020$ & $-21.958\pm0.019$ & $-13.372\pm 0.450$ & $151.12\pm0.28$ & $-1375.2 \pm 1.4$ \\ \midrule[0em]
    		MADE & $-3.097\pm 0.030$ & $3.306\pm 0.039$ & $-21.804\pm0.020$ & $-15.635\pm 0.498$ & $146.37\pm 0.28$ & $-1380.8 \pm 4.8$\\
    		MADE MoG & {\color{black} $\mathbf{0.375\pm 0.013}$} & $7.803\pm 0.022$ & {\color{black}$\mathbf{-18.368\pm 0.019}$} & $-12.740\pm 0.439$ & {$150.84\pm0.27$} & {\color{black}$\mathbf{-1038.5 \pm 1.8}$}\\ \midrule[0em]
    		Real NVP (10) & $0.182\pm 0.014$ & {\color{black}$\mathbf{8.357\pm 0.019}$} & {$-18.938\pm 0.021$} & {\color{black} $\mathbf{-11.795\pm 0.453}$} & {\color{black} $\mathbf{153.28\pm 1.78}$} & $-1370.7 \pm 10.1$\\ %
    		Real NVP (5) & {$-0.459\pm 0.010$} & {$6.656\pm 0.020$} & $-20.037\pm 0.020$ & $-12.418\pm 0.456$ & $151.76\pm 0.27$ & $-1323.2 \pm 6.6$\\ \midrule[0em]
    		MAF (5) & $-0.458\pm 0.016$ & {$7.042\pm 0.024$} & {$-19.400 \pm 0.020$} & $-11.816\pm 0.444$ & $149.22\pm 0.28$ & $-1300.5 \pm 1.7$\\
    		MAF (10) & $-0.376\pm 0.017$ & {$7.549\pm 0.020$} & {$-25.701\pm 0.025$} & $-11.892\pm 0.459$ & {$150.46\pm 0.28$} & $-1313.1 \pm 2.0$\\
    		MAF MoG (5) & $0.192\pm 0.014$ & {$7.183\pm 0.020$} & {$-22.747\pm 0.017$} & $-11.995\pm 0.462$ & {$152.58\pm 0.66$} & $-1100.3 \pm 1.6$\\  \bottomrule
    	\end{tabular}
	}
\end{table}

%% file: app/backprop.tex
We will follow~\cite{rojas2013neural}, Chapter 7, section 7.3.3 for the notation.
Let us define a two-layer neural network
\begin{equation}
    \mathbf{g}_\theta(\mathbf{x})= \boldsymbol{\sigma}\left(\mathbf{W}_{2} \boldsymbol{\sigma}\left(\mathbf{W}_{1} \mathbf{x}\right)\right)
\end{equation}
where we also define
\begin{align*}
\zb_2&=\boldsymbol{\sigma}\left(\mathbf{W}_{2} \zb_1\right)\\ 
\zb_1&=\boldsymbol{\sigma}\left(\mathbf{W}_{1} \mathbf{x}\right)\,.
\end{align*}
and
\begin{align*}
\mathbf{u}_{2}&=\boldsymbol{\sigma}^\prime\left(\mathbf{W}_{2}\zb_1\right)\\
\mathbf{u}_{1}&=\boldsymbol{\sigma}^\prime(\mathbf{W}_{1}\mathbf{x})
\end{align*}
and
\begin{align*}
\mathbf{y}_{2}&=\mathbf{W}_{2}\zb_1\\
\mathbf{y}_{1}&=\mathbf{W}_{1}\mathbf{x}
\end{align*}
We need to consider the contributions to the objective function due to the terms $\Lcal_p$ and $\Lcal^1_J$ (the contribution due to $\Lcal^2_J$ will be dealt with separately).
For $\Lcal_p$, we define
$$
e(x) = \frac{\partial}{\partial x} \log p(x^\prime)|_{x^\prime=x}
$$
and
$$
\mathbf{e}=\left(\begin{array}{c}{e(z_2^1) } \\ {e(z_2^2)} \\ {\vdots} \\ {e(z_2^D)}\end{array}\right)
$$
To deal with the terms in $\Lcal^1_J$, we define
\begin{align}
h(x) &= \frac{\partial}{\partial x} \log  x^\prime|_{x^\prime=x} \\
&= \frac{1}{x}    
\end{align}
and
$$
\mathbf{h}_{k}=\left(\begin{array}{c}{h(u_k^{1}) } \\ {h(u_k^{2})} \\ {\vdots} \\ {h(u_k^{D})}\end{array}\right)
$$
for $k=1,2$.
During forward propagation, we store the $\mathbf{D}_{k} = \operatorname{diag}\left(\boldsymbol{\sigma}^{\prime}\left(\mathbf{y}_{k}\right)\right)$ for $k = 1, 2$,
$$
\mathbf{D}_{k}=\left(\begin{array}{cccc}
{\sigma^\prime(y_{k}^1) } & {0} & {\cdots} & {0} \\
{0} & {\sigma^\prime(y_{k}^2)} & {\cdots} & {0} \\
{\vdots} & {\vdots} & {\ddots} & {\vdots} \\
{0} & {0} & {\cdots} & {\sigma^\prime(y_{k}^D)}
\end{array}\right)
$$
and the $\mathbf{G}_{k} = \operatorname{diag}\left(\boldsymbol{\sigma}''\left(\mathbf{y}_{k}\right)\right)$ for $k = 1, 2$,
$$
\mathbf{G}_{k}=\left(\begin{array}{cccc}
{\sigma''(y_{k}^1) } & {0} & {\cdots} & {0} \\
{0} & {\sigma''(y_{k}^2)} & {\cdots} & {0} \\
{\vdots} & {\vdots} & {\ddots} & {\vdots} \\
{0} & {0} & {\cdots} & {\sigma''(y_{k}^D)}
\end{array}\right)
$$
for example, if the nonlinearity were a sigmoid function $\sigma(x) = (1 + e^{-x})^{-1}$, the second derivative would be $\sigma''(x) = \sigma(x)(1-\sigma(x))\left(1-2\sigma(x)\right)$.
Then
$$
\boldsymbol{\delta}_{2} = \mathbf{D}_{2} \mathbf{e} + \mathbf{G}_{2} \mathbf{h}_{2}
$$
and
$$
\boldsymbol{\delta}_{1} = \mathbf{D}_{1}\mathbf{W}_2 \boldsymbol{\delta}_{2} + \mathbf{G}_{1} \mathbf{h}_{1}
$$
In general, the following recursive relationship holds
\begin{equation}
\label{eq:delta_rec}
\boldsymbol{\delta}_{k} = \mathbf{D}_{k}\mathbf{W}_{k+1} \boldsymbol{\delta}_{k+1} + \mathbf{G}_{k} \mathbf{h}_{k}
\end{equation}
Which results in the update rule
$$\Delta \mathbf{W}_{k}=-\mu \mathbf{z}_{k-1} \boldsymbol{\delta}_{k}^{\top}\,,$$
where $\zb_0=\xb$.
Notice that the only necessary operations are vector-matrix, matrix-vector and vector-vector multiplications.
\subsection{Relative gradient}
Now if we want to use the relative/natural gradient trick each of these terms needs to be multiplied by $\mathbf{W}_{k}^{\top} \mathbf{W}_{k}$ from the right.
$$
\Delta \mathbf{W}_{k}=-\mu  \mathbf{z}_{k-1} \boldsymbol{\delta}_{k}^{\top} \mathbf{W}_{k}^{\top} \mathbf{W}_{k}\,.
$$
\paragraph{Terms in $\Lcal_J^2$}
The terms in $\Lcal_J^2$, consisting of $\log |\mathbf{W}_k|$ give as gradient $\left(\mathbf{W}_k^{\top}\right)^{-1}$.
This requires a $D \times D$ matrix inversion for each of the matrices.
Our strategy to avoid it is to substitute the ordinary gradient with a relative gradient, where we multiply the gradient (with respect to the whole objective but for each layer separately) by $\mathbf{W}_{k}^{\top} \mathbf{W}_{k}$ from the right. 
In this case, the updates for the $\mathbf{W}_{k}$ terms simply become proportional to the $\mathbf{W}_{k}$ themselves.
Therefore, the update rule becomes
\begin{equation}
\label{eq:grad_updates}
\Delta \mathbf{W}_{k}=-\mu(  \mathbf{z}_{k-1} \boldsymbol{\delta}_{k}^{\top} \mathbf{W}_{k}^{\top}\mathbf{W}_{k} + \mathbf{W}_{k})\,.
\end{equation}
As we already noted, the operations involved in these updates can be performed in a way such that no matrix-matrix multiplication needs to be performed -- only matrix-vector and vector-vector multiplication. This is more apparent when the update rules are rewritten as below
\begin{equation}
\Delta \mathbf{W}_{k}=-\mu \left( \zb_{k-1}\left( \left( \bm{\delta}_k^{\top} \Wb_k^{\top} \right)\mathbf{W}_{k}\right) + \mathbf{W}_{k} \right)\,.
\end{equation}

%% file: app/app_rel_work.tex
In the following, we present a review of related work in tractable deep density estimation and invertible neural networks.

\textbf{Normalizing flows}
The modern conception of normalizing flows was introduced in~\cite{tabak2013family}, which discussed density estimation through the composition of simple maps. In~\cite{rippel2013high}, it was then proposed that deep density models implemented through neural networks could be used in order to construct bijective maps to a representation space and obtain normalized probability density estimates. Since then, the focus mainly shifted to scalability; \cite{dinh2014nice, dinh2016density} introduced scalable architectures, further refined in~\cite{kingma2018glow} to make them more scalable and suitable for practical applications; ~\cite{rezende2015variational} applied the results to variational inference. Comprehensive reviews on normalizing flows can be found in~\cite{papamakarios2019normalizing, kobyzev2019normalizing}. 

\textbf{Autoregressive flows}
Autoregressive flows are among the most used in practice. They involve maps which can be written as
$\mathrm{z}_{i}^{\prime}=\tau(\mathrm{z}_{i} ; \boldsymbol{h}_{i})$, with \hbox{$ \boldsymbol{h}_{i}=c_{i}\left(\mathbf{z}_{<i}\right)$}. $\tau$ is termed the \emph{transformer} and is a strictly monotonic function of $\mathrm{z}_{i}$, and $c_i$ is termed the $i$-th \emph{conditioner}. Its constraint is that the $i-$th conditioner can only take variables with dimension indices less than $i$ as an input. This results in an overall transformation with a triangular Jacobian; the determinant is therefore tractable and can be computed in $\mathcal{O}(D)$ time. Autoregressive flows differ in the way the transformer and conditioner are implemented; most commonly used are affine autoregressive flows \cite{dinh2014nice, dinh2016density, kingma2016improved, papamakarios2017masked, kingma2018glow} and non-affine neural transformers \cite{huang2018neural}.

\textbf{Linear flows}
A strict generalization of autoregressive flows, where the Jacobian is not constrained to be triangular, is given by linear flows, which are essentially transformations of the form $\zb' = \Wb \zb$, where $\Wb$ is a $D \times D$ invertible matrix. The Jacobian of the trasformation is simply $\Wb$ and both computing and optimizing its determinant takes time $O(D^3)$ in general. To obtain a better scaling behaviour, \cite{dinh2014nice} and~\cite{hoogeboom2019emerging} proposed to parameterize the invertible $\Wb$ matrix via matrix decomposition. One possibility is to compute the $\Pb\Lb\Ub$ decomposition of $\Wb$ and optimize the $\Lb$ and $\Ub$ triangular transformations. The drawback in this approach is that the permutation matrix $\Pb$ cannot be learned. A more flexible alternative is to consider the $\Qb\Rb$ decomposition of $\Wb$, where $\Qb$ is an orthogonal matrix and $\Rb$ is upper triangular. However computing $\Qb$ in full generality requires $O(D^3)$ operations, matching the complexity of the naive optimization of linear flows. \cite{tomczak2016improving} showed that we can apply the $\Qb$ transformation as a sequence of at most $D$ symmetry transformations each taking linear time, effectively making it possible to compute and optimize the $\Qb\Rb$ parameterization of $\Wb$ in $O(D^2)$ time; note however that the sequential nature of the computation makes the method unsuitable for optimization on hardware accelerators. An experimental comparison of the performance of the $\Pb\Lb\Ub$ and $\Qb\Rb$ decompositions against the direct optimization of $\Wb$ is found in~\cite{hoogeboom2019emerging}.%

\textbf{Flows based on residual transformations}
Another class of normalizing flows is based on invertible transformations of the form $\mathbf{z}^{\prime}=\mathbf{z}+g_{\phi}(\mathbf{z})$; this kind of flows are termed \emph{residual flows}. Two main approaches can be applied to build invertible residual flows: the first exploits the matrix determinant lemma and also results in determinants with $\mathcal{O}(D)$ computation time; however, there is no analytical way of computing their inverse. Examples of these approaches are Sylvester flows~\cite{berg2018sylvester}, planar flows~\cite{rezende2015variational} and radial flows~\cite{tabak2013family,rezende2015variational}. The second approach is that of contractive flows~\cite{behrmann2018invertible}: in this case, the determinant can not be computed simply; likelihood-based training of these models therefore needs to rely on a Hutchkinson's trace based approximation to the exact log-likelihood.

\textbf{Continuous time flows} A separate line of work focuses on building \emph{continuous flows}; in these approaches, the flow's infinitesimal dynamics is parametrized in continuous time, and the corresponding transformation is then found by integration~\cite{chen2018neural, grathwohl2018ffjord}; Hamiltonian Flows~\cite{rezende2015variational} can also be regarded as such kind of flows.

\textbf{Other works} Recently, many works have proposed ways of incorporating convolutional modules in normalizing flows, for example see~\cite{kingma2018glow, hoogeboom2019emerging, karami2019invertible}. In particular, ~\cite{finzi2019invertible} presents a formalization of the problem which bears some similarities to ours, while focusing on convolutional layers instead of fully connected ones.
Other work has been dedicated to constructing invertible neural networks, see for example \cite{baird2005one, jacobsen2018revnet, gomez2017reversible}.

%% file: app/complexity.tex
We want to characterize the complexity of computing
\begin{equation}
\nabla_{\mathbb{\boldsymbol{\theta}} } \log |\det \Jb\gb_{\boldsymbol{\mathbb{\theta}}}(\xb)|\,,     \label{eq:exact_}
\end{equation}
where $\gb_{\boldsymbol{\mathbb{\theta}}}$ is a neural network. 

We will first recapitulate the computational complexity of the main mathematical operations we employ (see e.g.~\cite{wiki:Computational_complexity_of_mathematical_operations}). Then we'll recapitulate the complexity of forward evaluation and backpropagation in neural networks. 
Finally, we'll discuss the implications on the complexity of computing the term in equation~\eqref{eq:exact_} with the three methods discussed in the paper --- namely, based on automatic differentiation, the standard computation described in section~\ref{sec:rel_grad} and the relative gradient based computation.

\subsection{Matrix operations} 

\paragraph{Matrix-vector and vector-vector multiplication} The multiplication of a $D \times D$ matrix with a $D \times 1$ vector scales as $\mathcal{O}(D^2)$. Same for the outer product between two vectors of dimension $D \times 1$.

\paragraph{Matrix-matrix multiplication} For the multiplication of two square matrices of size $D \times D$
\begin{itemize}
    \item An implementation of the Bareiss algorithm would scale as $\mathcal{O}(D^{3})$;
    \item An implementation of the Strassen algorithm would scale as$\mathcal{O}(D^{2.807\ldots})$ ;
    \item An implementation of the Coppersmith-Winograd algorithm would scale as$\mathcal{O}(D^{2.373\ldots})$ .
\end{itemize}

In practice, what is usually implemented in linear algebra libraries is some flavor of the Strassen algorithm (this is because the Coppersmith-Winograd algorithm, while having a more favorable asymptotic behaviour, is effectively slower if $D$ is not extremely high).

\paragraph{Matrix inversion} To find the inverse of a matrix of size $D \times D$
\begin{itemize}
    \item An implementation of Gauss-Jordan elimination would scale as $\mathcal{O}(D^{3})$;
    \item An implementation of the Strassen algorithm would scale as $\mathcal{O}(D^{2.807\ldots})$ ;
    \item An implementation of the Coppersmith-Winograd algorithm would scale as $\mathcal{O}(D^{2.373\ldots})$ .
\end{itemize}

\paragraph{Determinant} To find the determinant of a matrix of size $D \times D$
    \begin{itemize}
        \item An implementation of the Bareiss algorithm would scale as $\mathcal{O}(D^{3})$;
        \item Algorithms based on fast matrix multiplication scale as$\mathcal{O}(D^{2.373\ldots})$ .
\end{itemize}
    
For simplicity, in most of our considerations on complexity we assume that the computation of the determinant, the computation of the inverse and the multiplication of two square matrices have cubic cost. Notice that the cost of these operations always dominates over that of matrix-vector and vector-vector multiplication.

\subsection{Other operations involved in the Jacobian term computation}

Other operations turn out to be ininfluent on the overall computational complexity. Namely logarithms, absolute values, sums have no relevant effect in terms of asymptotic scaling, since their computational cost is dominated by that of the most expensive matrix operations listed above.

\subsection{Complexity of neural network operations}

\paragraph{Forward pass in a neural network} 
The complexity of the forward pass in a neural network depends on the neural network structure. 
For simplicity, we will consider fully connected Neural Networks, which due to their dense structure will provide an upper bound for the complexity of most of the nets used in practice.
Given an input vector, the forward pass is comprised of a sequential series of matrix-vector operations, plus elementwise operations on the resulting vector. The matrix-vector operations dominate the complexity; for an $L$ layer neural network, there are $L$ such operations.
Therefore, for data of dimensionality $D$, the complexity of a forward pass in a Neural Network for a single data sample is $\mathcal{O}(LD^2)$. 
\paragraph{Minibatching} 
The objectives should, in principle, be optimized on the full batch.
Stochastic optimization~\cite{bottou2010large} relies on the idea that the update steps in the optimization process can be performed on subsets of the whole training data, called minibatches.  
In practice these objectives will always be computed on minibatches, so the expected value must be substituted with its empirical estimate over a single minibatch. The minibatch size should in principle be considered when considering how the algorithm scales.
In the remainder, however, we will neglect this term, as minibatches used in practice are usually quite small.

\paragraph{Gradient computation}  On top of this, we also need to consider the gradient computation. Since the gradient is taken over the scalar loss function, this implies (through backpropagation or reverse mode differentiation) no increase in the asymptotic computational cost. We further elaborate on this in the next section.

\subsection{Computing the Jacobian with automatic differentiation}

\paragraph{Jacobian through automatic differentiation} Automatic differentiation~\cite{baydin2018automatic} includes two main operational modes: the forward mode and the backward mode.
Consider the computation of the Jacobian of a function $\mathbf{g}_\theta): \mathbb{R}^D \rightarrow  \mathbb{R}^d$.
The complexity of computing the Jacobian will depend on whether we use forward or reverse mode AD. 
This changes the complexity of the operation: 
\begin{itemize}
    \item forward mode requires $D \, c \, \text{ops}(\mathbf{g}_\theta)$ operations, where $D$ is the dimensionality of the data and $c$ is a constant, $c<6$ and typically $c \in [2,3]$ (see~\cite{griewank2008evaluating});
    \item reverse mode requires $d \, c \, \text{ops}(\mathbf{g}_\theta)$ operations.
\end{itemize}
In the case of dimensionality reduction, reverse mode differentiation (of which backpropagation represents an instance) is clearly more efficient. This is the case when the output of the function is scalar ($d=1$); thus, this explains our claim that gradients computation with backpropagation implies no increase in the asymptotic computational cost with respect to the forward pass alone.

For neural networks where all layers (including input and output) have the same size, both methods result in the same complexity. So in that case neither is better in terms of computational complexity --- though in practice it is known that reverse mode performs better~\cite{margossian2019review}.
For such neural networks (including those we consider) therefore, given that $\text{ops}(\mathbf{g}_\theta)$ is $\mathcal{O}(LD^2)$, the overall complexity of the Jacobian computation via automatic differentiation is $\mathcal{O}(LD^3)$.

The gradient of the objective can then be computed via backpropagation; however, the forward evaluation is what dominates the overall complexity.

\paragraph{Standard and relative gradient computations}

The evaluation of the two terms $\Lcal_p$ and $\Lcal_J^1$ requires a forward pass of the neural networks, thus scaling as $\mathcal{O}(LD^2)$. As we discussed, backpropagation to compute the gradient does not increase the overall cost.
For $\Lcal_J^2$, as we have shown, the gradient can be computed without need to actually evaluate the corresponding term (that is, side-stepping the determinant computation). However, the standard computation of the gradient still requires computing inverses of all the weight matrices, resulting in a cubic cost operation for each layer --- thus utimately in $\mathcal{O}(LD^3)$ cost.

When using the relative gradient, this inversion can be avoided, and computing the gradients of $\Lcal_J^2$ implies \emph{no additional costs}. The overall cost of the gradient computation is therefore simply $\mathcal{O}(LD^2)$.

%% file: app/implementation.tex
To efficiently optimize our objective (e.g. equation~\eqref{eq:basic_likelihood} in the main paper) we need to implement a variant of the backpropagation algorithm as detailed in appendix \ref{sec:mat_bp}. In particular, we need to compute the updates (equation~\eqref{eq:grad_upd} in the main paper) avoiding expensive matrix-matrix multiplications. This section is devoted to the description of an implementation strategy that takes advantage of Automatic Differentiation (AD), in order to have full flexibility in the definition of our model architectures and loss functions.  

Although all modern deep learning frameworks include automatic differentiation libraries, they implement the standard backpropagation algorithm. To implement our variant, we have two straightforward alternatives:

\begin{itemize}
    \item tweak some existing AD libraries to let us access the extra terms we need;
    \item implement our own AD library with the extra functionality we need.
\end{itemize}

The second alternative is easily excluded as we don't want to reinvent the wheel and the development effort would be too much. The first alternative is somewhat viable, but not future proof; we would be faced with the need to support our own modifications on top of the AD library we use.

We obviate to these problems with a little trick: we introduce in our architectures some dummy layers to accumulate the partial results that the standard backpropagation computes in the backward pass. This approach solves the previous problems by being:

\begin{itemize}
    \item universal: it can be easily implemented on top of whatever AD library that computes reverse-mode AD, without tweaking the internals of the library;
    \item efficient: the dummy layer operations are \(\mathcal{O}(1)\).
\end{itemize}

\subsection{The Accumulator layer}

To obtain the gradient updates \eqref{eq:grad_updates} we need to compute the \(\bm{\delta}\) terms \eqref{eq:delta_rec}. To better understand what these terms represent, we can consider a simple 2-layers "scalar" network, i.e. a network in which inputs, outputs and weights are scalar values:

\begin{align}
\label{eq:scalar_net}
f(x; \bm{w}) &= w_2 \sigma(w_1 x) \\ \nonumber
&= w_2 \sigma (y_1)\\ \nonumber
&= w_2 z_1 \\ \nonumber
&= y_2
\end{align}

where \(\bm{w}\) is the vector of scalar parameters, \(\sigma\) is the activation function of choice and

\[y_1 = w_1 x, \quad y_2 = w_2 z_1, \quad z_1 = \sigma (y_1)\,.\]

Given a loss function \(\Lcal\), the gradient of \(\Lcal\) with respect to \(w_1\) is easily computed with application of the chain rule

\begin{align}
\label{eq:chain}
    \frac{\partial \Lcal}{\partial w_1} &= \frac{\partial \Lcal}{\partial y_2} \frac{\partial y_2}{\partial z_1} \frac{\partial z_1}{\partial y_1} \frac{\partial y_1}{\partial w_1} \\ \nonumber
\end{align}

In this simple case, it is easy to isolate \(\delta\) in the gradient equation:
\begin{equation}
    \frac{\partial \Lcal}{\partial w_1} = \delta_1 \frac{\partial y_1}{\partial w_1}
\end{equation}

Reverse mode AD libraries necessarily compute all the partial derivatives in \eqref{eq:chain} and thus the \(\delta_1\) term we need. Unfortunately, the partial results are usually not accessible by the users. To access such terms without dealing with the internals of the AD libraries, we can introduce a parameterized function

\begin{equation}
\label{eq:acc_def}
a(x; \lambda) = x + \lambda \nonumber
\end{equation}

and redefine our scalar network as

\begin{equation}
    f(x; \bm{w}) = w_2 \sigma(a(y_1))
\end{equation}

The gradient with respect to \(w_1\) becomes

\begin{align}
\label{eq:grad_dummy}
    \frac{\partial \Lcal}{\partial w_1} &= \frac{\partial \Lcal}{\partial y_2} \frac{\partial y_2}{\partial z_1} \frac{\partial z_1}{\partial a} \frac{\partial a}{\partial y_1} \frac{\partial y_1}{\partial w_1} \\ \nonumber
\end{align}

The introduction of \(a\) is only a trick; in order not to modify the gradients nor the behaviour of the scalar network, we require

\begin{align}
    a(y_1) &= y_1 \\ \nonumber
    \frac{\partial z_1}{\partial a} &= \frac{\partial z_1}{\partial y_1} \\ \nonumber
    \frac{\partial a}{\partial y_1} &= 1 \nonumber
\end{align}

which is easily achieved by setting \(\lambda = 0\).

The benefit of introducing this accumulator layer \(a\) is that now we can ask the AD library to compute the gradients with respect to the dummy parameter \(\lambda\); it is easy to verify that

\begin{equation}
    \frac{\partial a}{\partial \lambda} = \delta_1
\end{equation}

thus making it possible to obtain the \(\delta\) terms we need to compute \eqref{eq:grad_updates}.

%% file: app/univ_approx.tex
Universal approximation for densities is a property often discussed in the context of autoregressive normalizing flows.
It can be shown, based on the proof of existence and non-uniqueness of solutions to the nonlinear ICA problem~\cite{hyvarinen1999nonlinear}, that any distribution can be mapped onto a factorized base distribution by an invertible function with triangular Jacobian, provided that the function class used for this mapping is large enough.
Normalizing flows with triangular Jacobians and a high number of parameters therefore have this approximation capacity (see e.g.~\cite{huang2018neural}).
However, they can obviously not represent all possible \emph{functions} --- but only those with triangular Jacobians. They can therefore not be used to learn proper inverse functions and perform useful feature extraction.

A more general notion of universal approximation is the one usually discussed in the neural network literature, that is --- universal approximation for functions. It has been shown that standard multilayer feedforward networks can approximate any continuous function to any degree of accuracy.
% %
For example,~\cite{leshno1993multilayer} proved that
a standard multilayer feedforward network with a locally bounded piecewise continuous activation function can approximate any continuous function to any degree of accuracy if and only if the network's activation function is not a polynomial. 
Biases also play a crucial role in this proof, as universal approximation capacity wouldn't be possible without.

While the proof above does not directly apply to our case, since it requires hidden layers with arbitrary width, we discuss how to incorporate biases in our training procedure in appendix~\ref{app:augm_matrix}, in order to increase the expressivity of our model.
We describe the nonlinearities we employed in appendix~\ref{app:experiments}.

%% file: app/augm_matrix.tex
In order to allow for the training of neural networks with biases, we present a heuristic based on the fact that affine transformations involving vector-matrix products plus biases can be represented as a single matrix operation through the formalism of the augmented matrix (see e.g.~\cite{rojas2013neural}).

Linear affine operations of the form $\yb = \Wb \xb + \bb$ can be represented via an augmented matrix as follows
\begin{equation}
\label{eq:augmented_mat}
\left[\begin{array}{l}\yb \\ 1\end{array}\right]=\left[\begin{array}{ccc|c} & \Wb & & \bb \\ 0 & \ldots & 0 & 1\end{array}\right]\left[\begin{array}{c}\xb \\ 1\end{array}\right] = \overline{\Wb} \left[\begin{array}{c}\xb \\ 1\end{array}\right] \,,
\end{equation}
where we refer to the matrix $\overline{\Wb}$ as \emph{augmented matrix}.

The question is whether the relative gradient trick can be applied to the augmented matrix.
The main issue is that we would like, throughout our optimization procedure, to remain on the manifold of augmented matrices; that is, we do not want to change the last row of $\overline{\Wb}_k$.
Therefore, the problem becomes a constrained optimization problem. 

\textbf{The $\Lcal_J^2$ term} It is easy to see that $\operatorname{det} \overline{\Wb}_k = \operatorname{det} \Wb_k$. The ordinary gradient for all terms in the last column and row of $\overline{\Wb}_k$ will therefore be equal to zero, and this will not be changed by the relative gradient trick; therefore, the contribution of this term will not lead us away from the  manifold of augmented matrices.

\textbf{The $\Lcal_p$ and $\Lcal_J^1$ terms} Both the $\yb_k$ and $\zb_k$ terms will however be influenced by the presence of biases, so the gradients on the first $D$ elements of the last column (that is $\bb_k$) will be nonzero. Through the multiplication with $\overline{\Wb}_k^{\top}\overline{\Wb}_k$, the updates given by the relative gradient on the last row of $\overline{\Wb}_k$ will therefore in general be nonzero, thus implying moving outside of the manifold we are interested in.

To solve this issue, we use a projected gradient algorithm, enforcing that the update for the last row of $\overline{\Wb}_k$ at each step is equal to zero. We call this algorithm \emph{projected relative gradient descent}.

In practice, we can use the augmented matrix formalism to apply the relative trick to the full parameters and then extract only the updates for the parameters of interest \(\Wb\), \(\bb\) disregarding the dummy row in \eqref{eq:augmented_mat}. Denoting by \(\Gb\) the gradients of \(\Wb\) and by \(\gb_b\) the gradients of \(\bb\), we can compute the relative gradients as

\begin{equation}
    \left[\begin{array}{c|c} \Gb &  \gb_b \\ \hline \gb & g \end{array}\right] \overline{\Wb}^{\top}\overline{\Wb} = \left[\begin{array}{c|c} \Gb \Wb^{\top} \Wb + \gb_b \bb^{\top} \Wb &  \Gb \Wb^{\top} \bb + \gb_b \bb^{\top} \bb + \gb_b \\ \hline \dots & \dots \end{array}\right]
\end{equation}

The relative gradient updates we need are then given by

\begin{align}
    \Delta\Wb &\rightarrow \Gb \Wb^{\top} \Wb + \gb_b \left( \bb^{\top} \Wb \right) \\
    \Delta\bb &\rightarrow  \Gb \left( \Wb^{\top} \bb \right) + \gb_b (1 + \bb^{\top} \bb)
    \label{eq:bias_upd}
\end{align}

Note that \(\Gb\) is nothing more then the standard backpropagation update \eqref{eq:grad_upda}, thus we can efficiently compute \(\Delta\Wb\) by avoiding matrix-matrix multiplications as in \eqref{eq:grad_upd}. For \(\Delta\bb\) we can directly avoid matrix-matrix multiplications by taking some care in the evaluation of \eqref{eq:bias_upd}.
% 

% 
% 

% \clearpage

%% file: app/convolutions.tex
The convolutional neural network~\cite{wu2017introduction} is composed of modules whose main components are: (i) a convolution layer; (ii) a pooling layer; (iii) a nonlinearity.

\paragraph{The convolution operation}
We follow the same notation as in~\cite{wu2017introduction}.
Typically, inputs to the convolution layers are order $3$ tensors with size $H^l \times W^l \times D^l$.
A convolution kernel is also an order 3 tensor with size $H \times W^l \times D^l$. If $D$ convolutions are used, this results in a order 4 tensor $\mathbb{R}^{H \times W^l \times D^l \times D}$ of parameters. If the input is $H \times W^l \times D^l$ and the kernel size
is $H \times W^l \times D^l \times D$, the convolution result has size $(H^l - H+1) \times (W^l - W +1) \times D$.
In our setting, note that the number of channels which can be used in practice is constrained, due to the formula in equation~\eqref{eq:basic_likelihood}, which requires the input and output dimensionalities to be equal.

\paragraph{Are convolutional neural networks invertible?}
The convolution operation was shown to be invertible under some mild conditions. See~\cite{ma2018invertibility} and~\cite{finzi2019invertible}, section 3.3, describing how Gaussian (or Uniform) sampled $c\times c\times r\times r$ parameter tensors will yield invertible convolutional layers with probability $1$.

The pooling layer can be substituted with an invertible counterpart (see~\cite{jacobsen2018revnet}, section 3; or~\cite{finzi2019invertible}, figure 3), which basically becomes a tensorial extension of the permutation operation.
As usual, an invertible nonlinearity can be chosen. 

\paragraph{Relative gradient for the convolution} For a convolution layer that preserves the number of channels in the input, we can directly write the operation in the form of a square matrix. In this case we can compute the relative gradient as explained in section~\ref{sec:derivation}, and we can obtain the gradients with respect to the filter entries by careful application of the chain rule.
We however leave the precise theoretical derivation and experiments for future work. 

\clearpage

%% file: app/computation_rel_vs_ord.tex
\label{app:exp_computaion}
\begin{figure}
\center
    \includegraphics{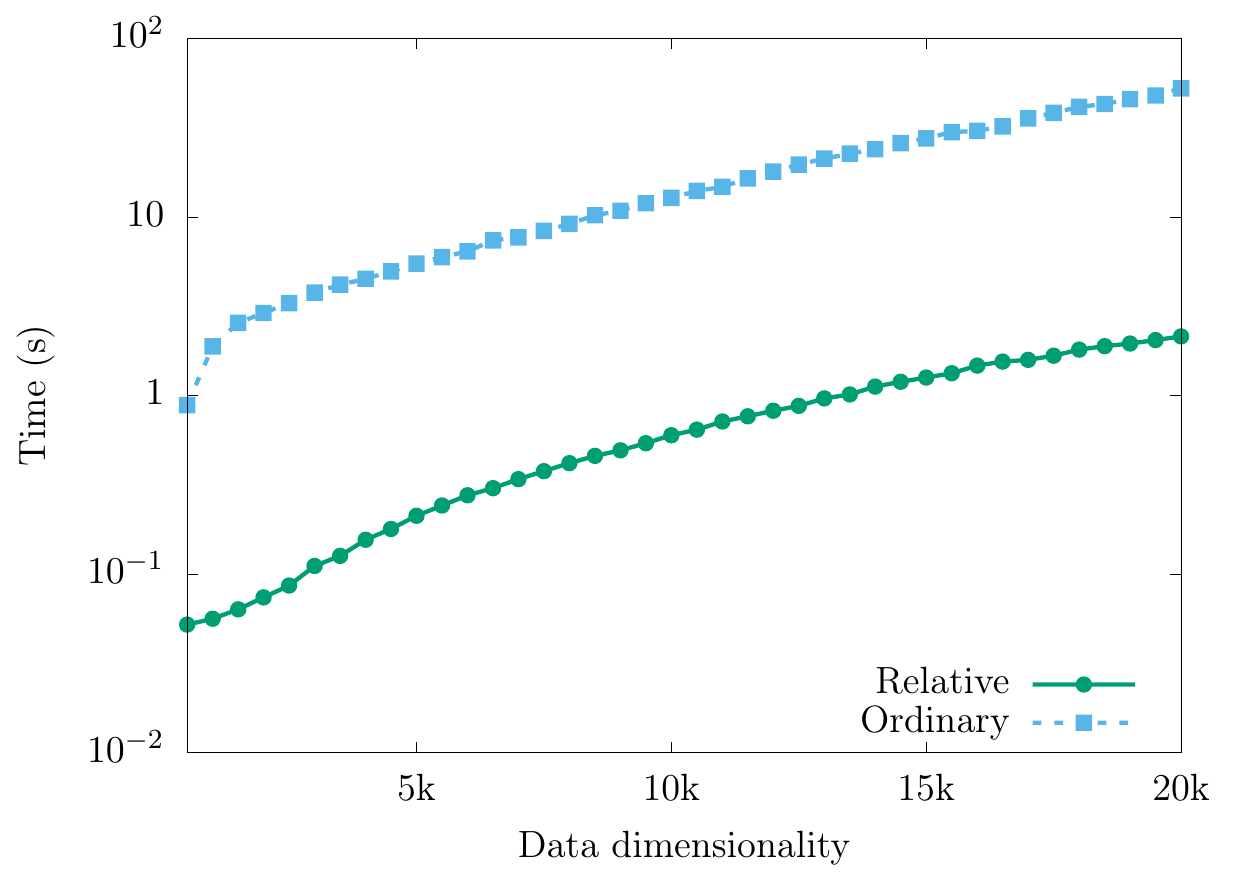}
    \caption{
    Comparison of the average computation times of a single evaluation of the gradient of the log-likelihood over a batch of size 100. Values are the mean over 5 steps, and the experiments have been run 5 times on a CPU cluster.}
    \label{fig:complexity_cpu}
\end{figure}

\textbf{Computational cost}
In section~\ref{sec:expts} and figure~\ref{fig:comparison}
we compared the computational cost of computing log-likelihood gradients with our newly proposed method and a naive backpropagation implementation when using hardware accelerators. Specifically, we used one Tesla P100 GPU card equipped with 16 GB of dedicated memory and circa 3500 computing cores. In figure \ref{fig:complexity_cpu} we show the same comparison for a computation platform comprising 48 cpu threads (Intel Xeon Processor E5-2650 v4 @ 2.20 GHz base frequency, 2.90 GHz max frequency) operating in parallel with about 250 GB of available RAM memory. It is hard to spot the expected theoretical improvement from $O(D^3)$ to $O(D^2)$, but a practical gain of about 2 orders of magnitude in computation time emerges in favor of the relative gradient computation.

In order to directly compare the execution times disregarding bottlenecks due to memory operations, we performed all of the experiments with no garbage collection. Anyways, by using always the same batch we made our experiments not very memory intensive and repeating the experiments with garbage collection enabled didn't show any substantial difference; we therefore don't report the plot.

\textbf{Memory consumption}
It is usual in deep learning to be constrained by the memory consumption of the models in use, as the available memory on hardware accelerators is typically scarce. To operate, a network needs to store the data, the intermediate activations (needed to compute gradients) and the parameters. For our simple architecture, the bottleneck is the storage of the parameters; this is because we don't employ very deep architectures, so the amount of intermediate activations to store is limited, and the size of the parameters grows quadratically with respect to the data size, meaning that parameters storage clearly dominate over data storage (this is assuming that data are loaded in small minibatches, which is the norm). This is certainly problematic for very high-dimensional datasets (i.e. high definition images) but even from this point of view we have a clear advantage over an explicit optimization of the Jacobian term with automatic differentiation. In this latter case, in fact, we need to compute the full Jacobian of the affine transformations for each individual data point; like for the weight matrices, the size of these terms grows quadratically with the input size, further increasing the memory footprint of the optimization procedure. 

As a simple example, we can compare the approximate memory requirements of the two methods in the moderately high-dimensional case with $D = 20000$. For a modest 2-layers network and employing Float32 weights (each requiring 4 Bytes (B) for storage), the memory needed to store the parameters amounts to $D^2 \times 4 B \times 2 \text{(layers)} = 3.2 GB$. Assuming a minibatch size of 100, data and activations require around 10-100 MB which is clearly negligible. The computed gradients will require the same space as the parameters, raising the memory footprint to over 6GB. For the gradient computations themselves, our method doesn't require additional memory (theoretically), while explicit automatic differentiation requires storing as many jacobian terms as the size of the minibatch, thus requiring over 300GB in our simple case. As this is clearly unfeasible on common hardware accelerators, we can drop the parallelization of the jacobian terms computation to considerably reduce memory consumption (bringing it down to over 9GB in our case), but this comes at the cost of further slowing down an already inefficient procedure.

\begin{figure}
  \centering
    \includegraphics{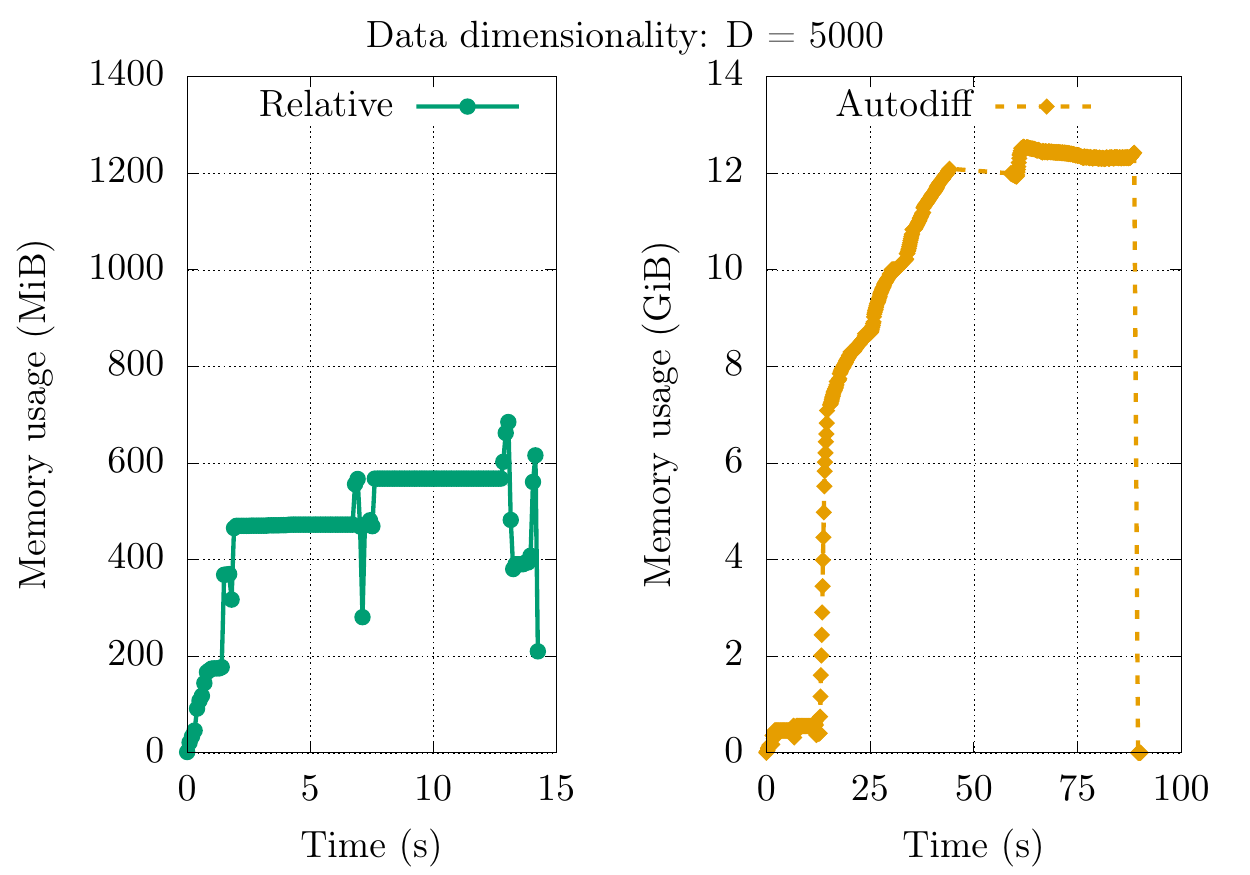}
    \caption{Comparison of the memory consumption for a single gradient evaluation. With D = 5000 our simplified analysis predicts a lower bound in the memory consumption of 400 MB for storing the parameters and the computed gradients; given that at startup time we observe a base memory consumption of almost 200 MB (computing environment + loaded libraries) we can see that our relative gradient implementation comes very close to the theoretical limit. For the naive autodiff implementation, instead, we compute a lower bound of 10.4 GB, which is approximately reflected in the empirical measurements (maximum consumption is almost 13 GB). Note: memory consumption for the autodiff case is reported in GiB, effectively making the scale of the plot one order of magnitude higher then in the relative gradient plot.}
    \label{fig:memory}
\end{figure}

While the simple analysis above shows a clear advantage for our proposed method, from the practical point of view many additional technical details might play a role in incrementing the memory requirements of both methods (e.g. loading of libraries and computing environment, just-in-time compilation steps, intermediate computations that can't be fused together...). In figure~\ref{fig:memory} we report a simple profiling of the memory consumption of the two methods, which shows how the difference is relevant in practice.

%% file: app/optimization_rel.tex
In this section we report some additional observations analyizing the relative gradient optimization behaviour with different optimizers.

\begin{figure}[t]
    \centering
        \begin{minipage}{.164\textwidth}
            \begin{subfigure}{1.0\textwidth}
            \centering
            \includegraphics[height=2.29cm, keepaspectratio]{figures/toy_density_single/trimodal-original.png}
            \end{subfigure}
        \end{minipage}%
        \begin{minipage}{.164\textwidth}
            \begin{subfigure}{1.0\textwidth}
            \centering
            \includegraphics[height=2.29cm, keepaspectratio]{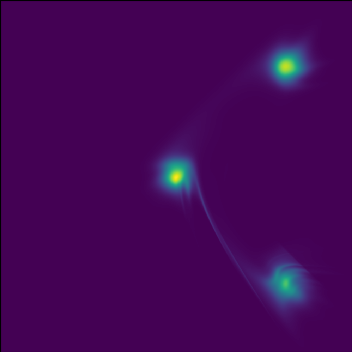}
            \end{subfigure}
        \end{minipage}%
        \hspace{.05em}
        \begin{minipage}{.164\textwidth}
            \begin{subfigure}{1.0\textwidth}
            \centering
            \includegraphics[height=2.29cm, keepaspectratio]{figures/toy_density_single/half-moons-original.png}
            \end{subfigure}
        \end{minipage}%
        \begin{minipage}{.164\textwidth}
            \begin{subfigure}{1.0\textwidth}
            \centering
            \includegraphics[height=2.29cm, keepaspectratio]{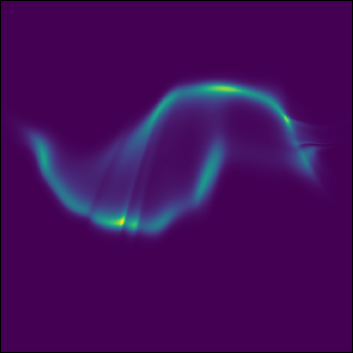}
            \end{subfigure}
        \end{minipage}%
        \hspace{.05em}
        \begin{minipage}{.164\textwidth}
            \begin{subfigure}{1.0\textwidth}
            \centering
            \includegraphics[height=2.29cm, keepaspectratio]{figures/toy_density_single/sine-original.png}
            \end{subfigure}
        \end{minipage}%
        \begin{minipage}{.164\textwidth}
            \begin{subfigure}{1.0\textwidth}
            \centering
            \includegraphics[height=2.29cm, keepaspectratio]{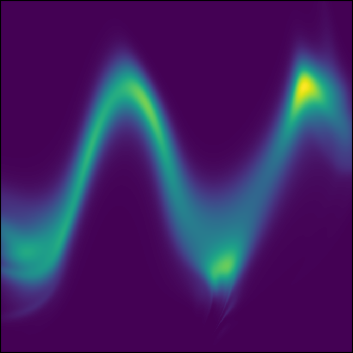}
            \end{subfigure}
        \end{minipage}%
        \caption{2D toy examples trained with SGD. True distribution on the left, predicted densities on the right.}
        \label{fig:toy_density_sgd}
\end{figure}

\begin{figure}
    \center
    \vspace{-2cm}
    \includegraphics[width=.9\linewidth]{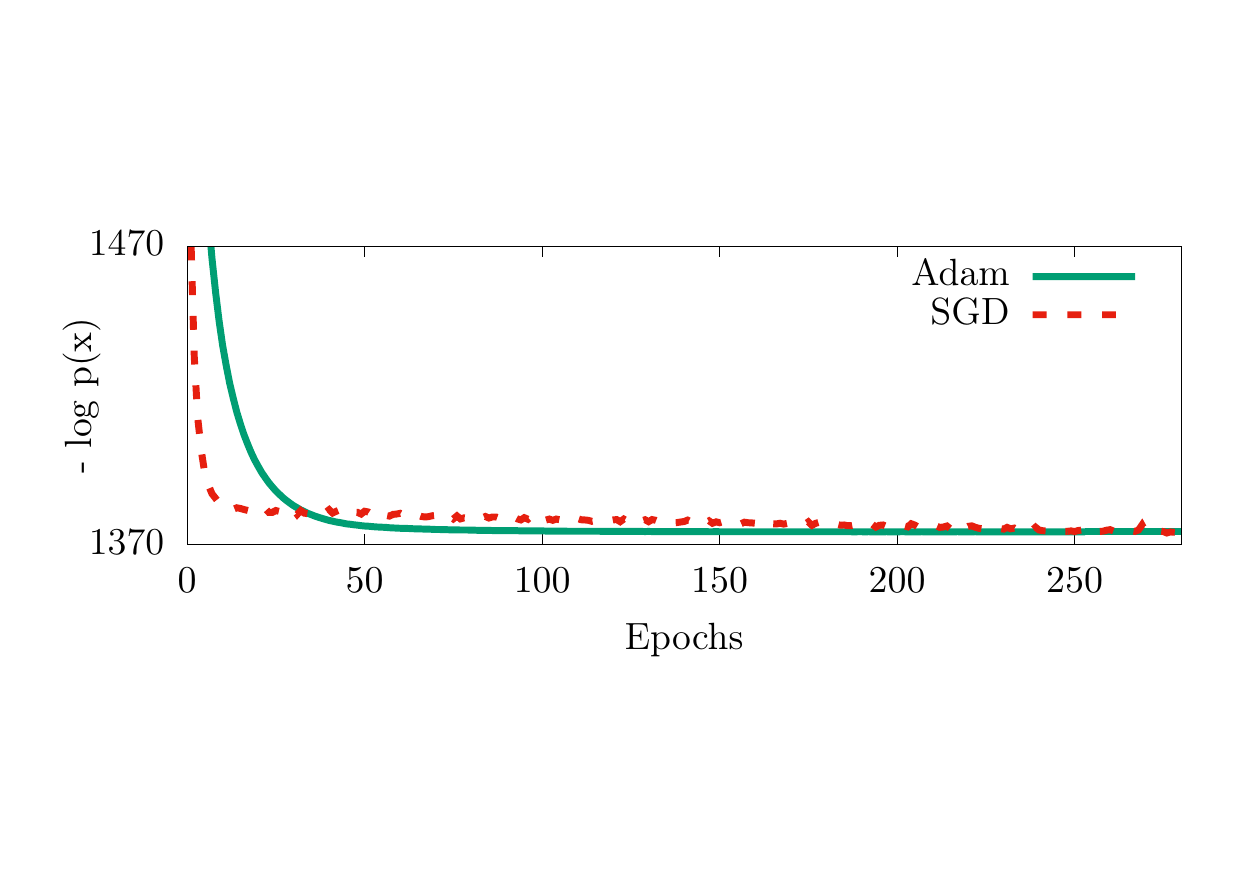}
    \vspace{-2cm}
    \caption{
    Log-likelihood evolution on MNIST validation set.}
    \label{fig:adam_vs_sgd}
\end{figure}

In figures \ref{fig:toy_density_sgd} and \ref{fig:adam_vs_sgd} we compare the optimization behaviour using vanilla Stochastic Gradient Descent (SGD) and Adam. Results on toy datasets like those in figure 2 in the main paper are shown in figure~\ref{fig:toy_density_sgd}. It can be seen that the data densities are modeled convincingly. We also report (figure~\ref{fig:adam_vs_sgd}) the evolution of the loss with SGD and Adam on density estimation on MNIST. % 
The two methods seem to reach convergence at comparable speed: SGD is faster initially, but in the longer run Adam appears to achieve a better performance faster.
Ultimately, both methods achieve a comparably good result. %

%% file: app/density.tex
\label{app:exp_density_est}
\paragraph{Architecture}
Although mentioned all throughout the paper, let us recall the neural network used for these experiments.
We here rely on the usual feedforward architecture, that is, a neural network for which the input is sequentially passed through an interleaving series of matrix multiplications and non-linear activation functions, being the last operation a matrix multiplication.
\paragraph{Nonlinearities}
Note that, since we make use of square weight matrices, the only two hyperparameters left in our architecture are the number of layers in the network, $L$, and the non-linearity used.
We consider two types of non-linearities. First, a smooth version of the leaky-ReLU activation function with a hyperparameter $\alpha$, 
\begin{equation}
    \operatorname{s_L}(x) = \alpha x + (1 - \alpha) \log (1 + e^x).
    \label{eq:sl_relu}
\end{equation}
Second, a weighted sum of the identity and hyperbolic tangent functions with two hyperparameters, $\alpha$ and $\beta$, controlling the steepness and ``level of linearity'' of the activation function,
\begin{equation}
    \operatorname{s_T}(x) = \tanh(\alpha x) + \beta x.
\end{equation}
However, in our experiments, these two hyperparameters for the $s_T$ nonlinearity are fixed to $\alpha=1$ and $\beta=0.1$ always.
Both of these nonlinearities are relatively smooth, and while no closed form solution for their inverse is available they can be inverted easily with a Newton method; in practice, for our parameter choice, we use a fixed number of $100$ iterations which seems to be (way) more than sufficient.

\paragraph{Toy examples}
For all the experiments shown in figure~\ref{fig:toy_density} of the main paper, we always use Adam as optimizer, fix the batch size and number of layers $L$ to $100$, use biases, and fix the activation function to $\operatorname{s_L}$ with $\alpha = 0.3$.
We chose as base distribution (that is, the distribution of the latent variables) the standard normal distribution.
We plot, as in the quantitative experiments, the best model found during the training.
Regarding the data, we sampled five-thousand samples for the training set and five-hundred points for the test set.
The only changing hyperparameters across the figures is the learning rate and the number of epochs, which are summarised in table~\ref{tab:toys}.

\begin{table}[h]
	\centering
	\caption{Hyperparameters used for figure 2 of the main paper.} \label{tab:toys}
% 	\resizebox{\textwidth}{!}{
    	\begin{tabular}{lccc} \toprule
    		\textbf{} & MoG & half moons & sine \\ \midrule
    		learning rate & $0.001$ & $0.001$ & $0.005$  \\ 
    		no. of epochs & $2000$ & $1300$ & $4000$ \\ \bottomrule
    	\end{tabular}
% 	}
\end{table}

\paragraph{Quantitative results on MNIST}
To obtain the density results on the MNIST dataset, the same preprocessing as in \cite{papamakarios2017masked} has been applied. For the model architecture, we fixed the number of layers to 2, we used the smooth Leaky-ReLU \eqref{eq:sl_relu} with $\alpha = 0.01$ and a standard normal distribution as a distribution for the latent variables. The optimization has been performed with Adam with default parameters. The hyperparameters search has been performed over learning rate values of $0.001, 0.0005, 0.0001$ and batch sizes of $10, 100$. For each run, we selected the model whose performance did not improve in the successive 30 epochs of training (i.e. we chose the model at epoch 10 if all the values of the loss for epochs 11 to 40 were higher then the value after 10 epochs). The best hyperparameters selection is shown in table~\ref{tab:best}.

\paragraph{Convergence time on MNIST} To get an idea of the running time of our method in a real-world scenario, one epoch on MNIST ($D = 784$, 50k training samples) on a modern laptop CPU takes an order of tens of seconds, a $\sim 4.5 \times$ speedup compared to ``standard'' optimization (which is roughly consistent with figure \ref{fig:complexity_cpu}, which was obtained with a slightly different experimental setup) and $\sim 50 \times$ speedup with respect to ``autodiff''. Our convergence time is $\sim 15$ min. While the speed-up is already visible at this data dimensionality, the difference is expected to be larger at higher dimensionality.

\paragraph{Quantitative results}
First, we want to remark that the data used for the experiments shown in table~\ref{tab:results} was pre-processed in the exact same way as described in~\cite{papamakarios2017masked}.

For the results shown in such table (MNIST excluded) a more exhaustive hyperparameter search has been performed. 
Particularly, for each dataset a grid-search was run with the options shown in table~\ref{tab:hyperparams}, taking for each experiment the model with best validation log-likelihood obtained during training and, across experiments, getting the one with best test log-likelihood.
Experiments were again trained using Adam and, instead of fixing the number of epochs, training was finished with an early-stopping criteria that evaluates the validation set every $25$ epochs and has a patience of $5$ trials.
The best hyperparameters selection is shown in table~\ref{tab:best}.

\begin{table}[ht]
	\centering
	\caption{Hyperparameters considered for the grid search.} \label{tab:hyperparams}
    	\begin{tabular}{lccc} \toprule
    		\textbf{} & Option~\#1 & Option~\#2 & Option~\#3 \\ \midrule
    		activation & $\operatorname{s_L}, \alpha=0.3$ & $\operatorname{s_L}, \alpha=0.01$ & $\operatorname{s_T}$  \\ 
    		no. layers & $25$ & $50$ & $100$ \\
    		learning rate & $0.001$ & $0.0005$ & $0.0001$ \\ 
    		batch size & $10$ & $50$ & $100$ \\
    		base distribution & standard normal & hyperbolic secant & \\
    		bias & Yes & No &  \\ 
    		\bottomrule
    	\end{tabular}
\end{table}

\begin{table}[ht]
	\centering
	\caption{Hyperparameters for the results in table 1 in the main paper.} \label{tab:best}
	\resizebox{\textwidth}{!}{
    	\begin{tabular}{lcccccc} \toprule
    		\textbf{} & POWER & GAS & HEPMASS & MINIBOONE & BSDS300 & MNIST \\ \midrule
    		activation & $\operatorname{s_L}, \alpha=0.3$ & $\operatorname{s_L}, \alpha=0.3$ & $\operatorname{s_L}, \alpha=0.3$ & $\operatorname{s_T}$ & $\operatorname{s_T}$ & $\operatorname{s_L}, \alpha=0.01$\\ 
    		no. layers & $50$ & $100$ & $50$ & $25$ & $25$ & $2$ \\
    		learning rate & $0.001$ & $0.001$ & $0.001$ & $0.0001$ & $0.0001$ & $0.0001$ \\ 
    		batch size & $100$ & $100$ & $50$ & $100$ & $100$ & $10$\\
    		base dist. & std normal & std normal & hyper. secant & std normal & hyper. secant & std normal \\
    		bias & Yes & Yes & No & Yes & No & Yes \\ 
    		\bottomrule
    	\end{tabular}
	}
\end{table}

Regarding the rest of the models shown in that table, we reproduce the exact same experiments as those described in~\cite{papamakarios2017masked}.
Therefore, the considered models have the same architecture and stopping criteria as the ones shown in table~1 of the aforementioned paper.
The only difference with respect to the results shown in table~1 of \cite{papamakarios2017masked} and table~\ref{tab:results} in our paper is the number of trainable parameters.        
As mentioned in section~\ref{sec:expts}, in order to perform a fair comparison, we tweaked the hyperparameters of each architecture so they have approximately the same number of parameters.

Specifically, we first trained our model as described above and, once we knew the number of parameters of the best-performing model (which is approximately $LD^2$) we used the formulae shown in table~3 of \cite{papamakarios2017masked} to find to which values we should fix the hyperparameters $L$ and $H$ of their models so that they have the same number of parameters.

As a final remark, note that there is one degree-of-freedom in those equations (for every $L$ there is a value of $H$ solving the given equation). Therefore, for each of the considered models and datasets, we run two different experiments, one with $L=1$ and another with $L=2$ (as similarly done in~\cite{papamakarios2017masked}), finding afterwards the proper value of $H$ to match the number of trainable parameters of our best model for that same dataset.

%% file: log_det_jac.bbl
\begin{thebibliography}{10}

\bibitem{absil2009optimization}
P-A Absil, Robert Mahony, and Rodolphe Sepulchre.
\newblock {\em Optimization algorithms on matrix manifolds}.
\newblock Princeton University Press, 2009.

\bibitem{amari1998natural}
Shun-Ichi Amari.
\newblock Natural gradient works efficiently in learning.
\newblock {\em Neural computation}, 10(2):251--276, 1998.

\bibitem{baird2005one}
Leemon Baird, David Smalenberger, and Shawn Ingkiriwang.
\newblock One-step neural network inversion with pdf learning and emulation.
\newblock In {\em Proceedings. 2005 IEEE International Joint Conference on
  Neural Networks, 2005.}, volume~2, pages 966--971. IEEE, 2005.

\bibitem{baydin2018automatic}
Atilim~Gunes Baydin, Barak~A Pearlmutter, Alexey~Andreyevich Radul, and
  Jeffrey~Mark Siskind.
\newblock Automatic differentiation in machine learning: a survey.
\newblock {\em Journal of machine learning research}, 18(153), 2018.

\bibitem{behrmann2018invertible}
Jens Behrmann, Will Grathwohl, Ricky~TQ Chen, David Duvenaud, and
  J{\"o}rn-Henrik Jacobsen.
\newblock Invertible residual networks.
\newblock In {\em International Conference on Machine Learning}, pages
  573--582, 2019.

\bibitem{bell1995information}
Anthony~J Bell and Terrence~J Sejnowski.
\newblock An information-maximization approach to blind separation and blind
  deconvolution.
\newblock {\em Neural computation}, 7(6):1129--1159, 1995.

\bibitem{blei2017variational}
David~M Blei, Alp Kucukelbir, and Jon~D McAuliffe.
\newblock Variational inference: A review for statisticians.
\newblock {\em Journal of the American statistical Association},
  112(518):859--877, 2017.

\bibitem{bonnabel2013stochastic}
Silvere Bonnabel.
\newblock Stochastic gradient descent on riemannian manifolds.
\newblock {\em IEEE Transactions on Automatic Control}, 58(9):2217--2229, 2013.

\bibitem{bottou2010large}
L{\'e}on Bottou.
\newblock Large-scale machine learning with stochastic gradient descent.
\newblock In {\em Proceedings of COMPSTAT'2010}, pages 177--186. Springer,
  2010.

\bibitem{jax2018github}
James Bradbury, Roy Frostig, Peter Hawkins, Matthew~James Johnson, Chris Leary,
  Dougal Maclaurin, and Skye Wanderman-Milne.
\newblock {JAX}: composable transformations of {P}ython+{N}um{P}y programs,
  2018.

\bibitem{cardoso1996equivariant}
J-F Cardoso and Beate~H Laheld.
\newblock Equivariant adaptive source separation.
\newblock {\em IEEE Transactions on signal processing}, 44(12):3017--3030,
  1996.

\bibitem{chen2019neural}
Tian~Qi Chen and David~K Duvenaud.
\newblock Neural networks with cheap differential operators.
\newblock In {\em Advances in Neural Information Processing Systems}, pages
  9961--9971, 2019.

\bibitem{chen2018neural}
Tian~Qi Chen, Yulia Rubanova, Jesse Bettencourt, and David~K Duvenaud.
\newblock Neural ordinary differential equations.
\newblock In {\em Advances in Neural Information Processing Systems}, pages
  6572--6583, 2018.

\bibitem{dinh2014nice}
Laurent Dinh, David Krueger, and Yoshua Bengio.
\newblock {NICE}: Non-linear independent components estimation.
\newblock {\em arXiv preprint arXiv:1410.8516}, 2014.

\bibitem{dinh2016density}
Laurent Dinh, Jascha Sohl-Dickstein, and Samy Bengio.
\newblock Density estimation using real {NVP}.
\newblock {\em arXiv preprint arXiv:1605.08803}, 2016.

\bibitem{Dua:2019}
Dheeru Dua and Casey Graff.
\newblock {UCI} machine learning repository, 2017.

\bibitem{finzi2019invertible}
Marc Finzi, Pavel Izmailov, Wesley Maddox, Polina Kirichenko, and Andrew~Gordon
  Wilson.
\newblock Invertible convolutional networks.
\newblock 2019.

\bibitem{gomez2017reversible}
Aidan~N Gomez, Mengye Ren, Raquel Urtasun, and Roger~B Grosse.
\newblock The reversible residual network: Backpropagation without storing
  activations.
\newblock In {\em Advances in neural information processing systems}, pages
  2214--2224, 2017.

\bibitem{grathwohl2018ffjord}
Will Grathwohl, Ricky~TQ Chen, Jesse Bettencourt, Ilya Sutskever, and David
  Duvenaud.
\newblock {FFJORD}: Free-form continuous dynamics for scalable reversible
  generative models.
\newblock {\em arXiv preprint arXiv:1810.01367}, 2018.

\bibitem{gresele2019incomplete}
Luigi Gresele, Paul~K. Rubenstein, Arash Mehrjou, Francesco Locatello, and
  Bernhard Sch{\"{o}}lkopf.
\newblock The {I}ncomplete {R}osetta {S}tone problem: Identifiability results
  for multi-view nonlinear {ICA}.
\newblock In Amir Globerson and Ricardo Silva, editors, {\em Proceedings of the
  Thirty-Fifth Conference on Uncertainty in Artificial Intelligence, {UAI}
  2019, Tel Aviv, Israel, July 22-25, 2019}, page~53. {AUAI} Press, 2019.

\bibitem{griewank2008evaluating}
Andreas Griewank and Andrea Walther.
\newblock {\em Evaluating derivatives: principles and techniques of algorithmic
  differentiation}, volume 105.
\newblock Siam, 2008.

\bibitem{hoogeboom2019emerging}
Emiel Hoogeboom, Rianne van~den Berg, and Max Welling.
\newblock Emerging convolutions for generative normalizing flows.
\newblock {\em arXiv preprint arXiv:1901.11137}, 2019.

\bibitem{hornik1989multilayer}
K.~Hornik, M.~Stinchcombe, and H.~White.
\newblock Multilayer feedforward networks are universal approximators.
\newblock {\em Neural Netw.}, 2(5):359–366, July 1989.

\bibitem{hornik1991approximation}
Kurt Hornik.
\newblock Approximation capabilities of multilayer feedforward networks.
\newblock {\em Neural networks}, 4(2):251--257, 1991.

\bibitem{huang2018neural}
Chin-Wei Huang, David Krueger, Alexandre Lacoste, and Aaron Courville.
\newblock Neural autoregressive flows.
\newblock {\em arXiv preprint arXiv:1804.00779}, 2018.

\bibitem{hyvarinen1999fast}
Aapo Hyvarinen.
\newblock Fast and robust fixed-point algorithms for independent component
  analysis.
\newblock {\em IEEE transactions on Neural Networks}, 10(3):626--634, 1999.

\bibitem{hyvarinen2016unsupervised}
Aapo Hyvarinen and Hiroshi Morioka.
\newblock Unsupervised feature extraction by time-contrastive learning and
  nonlinear {ICA}.
\newblock In {\em Advances in Neural Information Processing Systems}, pages
  3765--3773, 2016.

\bibitem{hyvarinen2017nonlinear}
Aapo Hyv{\"a}rinen and Hiroshi Morioka.
\newblock Nonlinear {ICA} of temporally dependent stationary sources.
\newblock volume~54. Proceedings of Machine Learning Research, 2017.

\bibitem{hyvarinen1999nonlinear}
Aapo Hyv{\"a}rinen and Petteri Pajunen.
\newblock Nonlinear independent component analysis: Existence and uniqueness
  results.
\newblock {\em Neural Networks}, 12(3):429--439, 1999.

\bibitem{hyvarinen2018nonlinear}
Aapo Hyv{\"a}rinen, Hiroaki Sasaki, and Richard~E Turner.
\newblock Nonlinear {ICA} using auxiliary variables and generalized contrastive
  learning.
\newblock {\em arXiv preprint arXiv:1805.08651}, 2018.

\bibitem{jacobsen2018revnet}
J{\"o}rn-Henrik Jacobsen, Arnold Smeulders, and Edouard Oyallon.
\newblock i-{R}ev{N}et: Deep invertible networks.
\newblock {\em arXiv preprint arXiv:1802.07088}, 2018.

\bibitem{karami2019invertible}
Mahdi Karami, Dale Schuurmans, Jascha Sohl-Dickstein, Laurent Dinh, and Daniel
  Duckworth.
\newblock Invertible convolutional flow.
\newblock In {\em Advances in Neural Information Processing Systems}, pages
  5636--5646, 2019.

\bibitem{kingma2018glow}
Durk~P Kingma and Prafulla Dhariwal.
\newblock Glow: Generative flow with invertible 1x1 convolutions.
\newblock In {\em Advances in Neural Information Processing Systems}, pages
  10215--10224, 2018.

\bibitem{kingma2016improved}
Durk~P Kingma, Tim Salimans, Rafal Jozefowicz, Xi~Chen, Ilya Sutskever, and Max
  Welling.
\newblock Improved variational inference with inverse autoregressive flow.
\newblock In {\em Advances in neural information processing systems}, pages
  4743--4751, 2016.

\bibitem{kobyzev2019normalizing}
Ivan Kobyzev, Simon Prince, and Marcus~A Brubaker.
\newblock Normalizing flows: Introduction and ideas.
\newblock {\em arXiv preprint arXiv:1908.09257}, 2019.

\bibitem{lecun1989backpropagation}
Yann LeCun, Bernhard Boser, John~S Denker, Donnie Henderson, Richard~E Howard,
  Wayne Hubbard, and Lawrence~D Jackel.
\newblock Backpropagation applied to handwritten zip code recognition.
\newblock {\em Neural computation}, 1(4):541--551, 1989.

\bibitem{lecun1998mnist}
Yann LeCun, Corinna Cortes, and Christopher~JC Burges.
\newblock The {MNIST} database of handwritten digits, 1998.
\newblock {\em URL http://yann. lecun. com/exdb/mnist}, 10:34, 1998.

\bibitem{leshno1993multilayer}
Moshe Leshno, Vladimir~Ya Lin, Allan Pinkus, and Shimon Schocken.
\newblock Multilayer feedforward networks with a nonpolynomial activation
  function can approximate any function.
\newblock {\em Neural networks}, 6(6):861--867, 1993.

\bibitem{ma2018invertibility}
Fangchang Ma, Ulas Ayaz, and Sertac Karaman.
\newblock Invertibility of convolutional generative networks from partial
  measurements.
\newblock In {\em Advances in Neural Information Processing Systems}, pages
  9628--9637, 2018.

\bibitem{margossian2019review}
Charles~C Margossian.
\newblock A review of automatic differentiation and its efficient
  implementation.
\newblock {\em Wiley Interdisciplinary Reviews: Data Mining and Knowledge
  Discovery}, 9(4):e1305, 2019.

\bibitem{martin2001database}
David Martin, Charless Fowlkes, Doron Tal, and Jitendra Malik.
\newblock A database of human segmented natural images and its application to
  evaluating segmentation algorithms and measuring ecological statistics.
\newblock In {\em Proceedings Eighth IEEE International Conference on Computer
  Vision. ICCV 2001}, volume~2, pages 416--423. IEEE, 2001.

\bibitem{papamakarios2019normalizing}
George Papamakarios, Eric Nalisnick, Danilo~Jimenez Rezende, Shakir Mohamed,
  and Balaji Lakshminarayanan.
\newblock Normalizing flows for probabilistic modeling and inference.
\newblock {\em arXiv preprint arXiv:1912.02762}, 2019.

\bibitem{papamakarios2017masked}
George Papamakarios, Theo Pavlakou, and Iain Murray.
\newblock Masked autoregressive flow for density estimation.
\newblock In {\em Advances in Neural Information Processing Systems}, pages
  2338--2347, 2017.

\bibitem{rezende2015variational}
Danilo Rezende and Shakir Mohamed.
\newblock Variational inference with normalizing flows.
\newblock In {\em International Conference on Machine Learning}, pages
  1530--1538, 2015.

\bibitem{rippel2013high}
Oren Rippel and Ryan~Prescott Adams.
\newblock High-dimensional probability estimation with deep density models.
\newblock {\em arXiv preprint arXiv:1302.5125}, 2013.

\bibitem{rojas2013neural}
Ra{\'u}l Rojas.
\newblock {\em Neural networks: a systematic introduction}.
\newblock Springer Science \& Business Media, 2013.

\bibitem{rumelhart1986learning}
David~E Rumelhart, Geoffrey~E Hinton, and Ronald~J Williams.
\newblock Learning representations by back-propagating errors.
\newblock {\em {N}ature}, 323(6088):533--536, 1986.

\bibitem{seeger2008large}
Matthias~W Seeger and Hannes Nickisch.
\newblock Large scale variational inference and experimental design for sparse
  generalized linear models.
\newblock {\em arXiv preprint arXiv:0810.0901}, 2008.

\bibitem{squartini2005new}
Stefano Squartini, Francesco Piazza, and Ali Shawker.
\newblock New {R}iemannian metrics for improvement of convergence speed in
  {ICA} based learning algorithms.
\newblock In {\em 2005 IEEE International Symposium on Circuits and Systems},
  pages 3603--3606. IEEE, 2005.

\bibitem{tabak2013family}
Esteban~G Tabak and Cristina~V Turner.
\newblock A family of nonparametric density estimation algorithms.
\newblock {\em Communications on Pure and Applied Mathematics}, 66(2):145--164,
  2013.

\bibitem{tabak2010density}
Esteban~G Tabak, Eric Vanden-Eijnden, et~al.
\newblock Density estimation by dual ascent of the log-likelihood.
\newblock {\em Communications in Mathematical Sciences}, 8(1):217--233, 2010.

\bibitem{tomczak2016improving}
Jakub~M Tomczak and Max Welling.
\newblock Improving variational auto-encoders using {H}ouseholder flow.
\newblock {\em arXiv preprint arXiv:1611.09630}, 2016.

\bibitem{berg2018sylvester}
Rianne Van Den~Berg, Leonard Hasenclever, Jakub~M Tomczak, and Max Welling.
\newblock Sylvester normalizing flows for variational inference.
\newblock In {\em 34th Conference on Uncertainty in Artificial Intelligence
  2018, UAI 2018}, pages 393--402. Association For Uncertainty in Artificial
  Intelligence (AUAI), 2018.

\bibitem{wainwright2008graphical}
Martin~J Wainwright and Michael~I Jordan.
\newblock Graphical models, exponential families, and variational inference.
\newblock {\em Foundations and Trends{\textregistered} in Machine Learning},
  1(1-2):1--305, 2008.

\bibitem{wiki:Computational_complexity_of_mathematical_operations}
Wikipedia.
\newblock {Computational complexity of mathematical operations} ---
  {W}ikipedia{,} the free encyclopedia.
\newblock
  \url{http://en.wikipedia.org/w/index.php?title=Computational\%20complexity\%20of\%20mathematical\%20operations&oldid=958179308},
  2020.
\newblock [Online; accessed 11-June-2020].

\bibitem{wu2017introduction}
Jianxin Wu.
\newblock Introduction to convolutional neural networks.
\newblock 2017.

\end{thebibliography}
